\documentclass[onecolumn]{autart}

\newtheorem{theorem}{Theorem}

\newtheorem{proposition}[theorem]{Proposition}

\newtheorem{lemma}[theorem]{Lemma}
\newtheorem{remark}[theorem]{Remark}

\newenvironment{proofthm}[1]{\noindent{\em Proof:}}{ \hfill $\Box$ \\}

\RequirePackage{amsmath, amssymb, graphicx,url,xspace,subfigure,braket,epstopdf}

\DeclareMathOperator*\argmin{arg\,min}

\usepackage{graphicx,subfigure}
\usepackage{epsfig} 
\usepackage{mathptmx} 
\usepackage{times} 
\usepackage{amsmath} 
\usepackage{amssymb}  
\usepackage{amsfonts}
\usepackage{euscript}
\usepackage{color}

\begin{document}

\begin{frontmatter}

\title{Mathematical foundations of stable RKHSs}

\author[First]{Mauro Bisiacco}

\address[First]{Department of Information  Engineering, University of Padova, Padova, Italy (e-mail: bisiacco@dei.unipd.it)}

\author[Second]{Gianluigi Pillonetto}

\address[Second]{Department of Information  Engineering, University of Padova, Padova, Italy (e-mail: giapi@dei.unipd.it)}

%


\maketitle
\begin{abstract}
Reproducing kernel Hilbert spaces (RKHSs) are key spaces for machine learning that are becoming popular also for linear system identification. In particular, the so-called stable RKHSs can be used to model absolutely summable impulse responses. 
In combination e.g. with 
regularized least squares they can then be used to reconstruct dynamic systems 
from input-output data. In this paper we provide new structural properties of stable RKHSs. 
The relation between stable kernels and other fundamental classes, like those containing absolutely summable or finite-trace
kernels, is elucidated. 
These insights are then brought into the feature space context.
First, it is proved that any stable kernel admits feature maps induced by a basis of orthogonal eigenvectors 
in $\ell_2$. The exact connection with classical system identification 
approaches that exploit such kind of functions to model impulse responses is also provided. 
 Then, the necessary and sufficient stability condition 
for RKHSs designed by formulating kernel eigenvectors and eigenvalues is obtained. 
Overall, our new results provide novel mathematical foundations of stable RKHSs with
impact on stability tests, impulse responses modeling and computational efficiency of 
regularized schemes for linear system identification.
%
\end{abstract}

\end{frontmatter}

\section{Introduction}\label{Sec1}

Reproducing kernel Hilbert spaces (RKHSs) are particular spaces of functions
in one-to-one correspondence with the class of positive semidefinite kernels.
While RKHS theory has been mainly developed in the fifties  \cite{Aronszajn50,Bergman50},
such spaces have found first important applications in the eighties in the context of
statistics and computer vision \cite{Bertero:1988,Wahba1990,Poggio90}.
They were then brought to the attention of 
machine learning community in \cite{Girosi:1997}. 
Since then, they have become a fundamental tool for 
function estimation. Estimators based on RKHSs 
include  smoothing splines  \cite{Wahba1990}, regularization networks
\cite{Poggio90} and support vector machines \cite{Drucker97,Vapnik98}.
Combinations with deep networks are also described in \cite{KBdeep2009,Mik2018}.\\

The importance of RKHSs for function estimation from sparse and noisy data arises from several facts. 
First, a RKHS $\mathcal{H}$ inherits its properties from the associated kernel $K$.
This is important for modeling purposes since all the expected function properties 
can be encoded in the kernel design. For instance, a regular kernel induces an RKHS of continuous functions
whose norm can be used as regularizer to penalize solutions with unphysical oscillations.
Indeed, the most important kernel-based estimators optimize objectives
containing a loss that accounts for
adherence to experimental data and the RKHS norm that restores well-posedness.
Another fundamental aspect is that the kernel can include in an implicit way a 
very large (possibly infinite) number
of basis functions, leading to very flexible and computable models.
This result has also connection with the following important 
mathematical fact. 
Let $\mathcal{X}$ be the regressor space, i.e. the domain of the functions $f: \mathcal{X} \rightarrow \mathbb{R}$ contained in 
$\mathcal{H}$.
Then, given any positive semidefinite kernel $K$, 
there always exists at least one 
inner-product space $\mathcal{F}$ and 
one feature map $\phi:\mathcal{X}\rightarrow \mathcal{F}$ 
such that\footnote{One explicit example is the RKHS map  $\phi_{\mathcal{H}}:\mathcal{X} \rightarrow \mathcal{H}$ such that
$\phi_{\mathcal{H}}(x) = K(x,\cdot).$
It always satisfies (\ref{EQ01}) in view of the so-called reproducing property \cite{Aronszajn50}.}
\begin{equation}\label{EQ01}
K(x,y) = \langle \phi(x), \phi(y) \rangle_{\mathcal{F}}, \qquad \phi:\mathcal{X}\rightarrow \mathcal{F}.
\end{equation}
Above, the components of $\phi(x)$ are the basis functions induced by the kernel, 
e.g. $\phi_1(x)=1,\phi_2(x)=x,\phi_3(x)=x^2,\ldots$ describe polynomial models. 
Kernels can thus define expressive spaces by (implicitly) mapping
the space of the regressors $\mathcal{X}$ 
into high-dimensional feature spaces where linear machines can be employed. 
Nonlinear algorithms can be reduced to linear ones
without even knowing explicitly the feature map.
In fact, under mild assumptions,
the kernel-based estimate is given by the sum of a finite number of kernel sections $K(x,\cdot)$
centred on the observed regressors \cite{Wahba1990,Scholkopf01}.

Control community's interest has been recently addressed to 
RKHSs tailored for linear system identification. 
While function models adopted in machine learning  
typically embed information e.g. on regularity, periodicity or sparsity, 
the new spaces hinge on kernels accounting for dynamic systems features.
Examples are the so-called stable spline, TC and DC kernels
that incorporate exponential stability 
\cite{PillACC2010,SS2010,ChenOL12,SSvsNN2013}, see also \cite{ChenKS2018,ChenetalTAC:13,PillonettoWien13,PillonettoHybrid} for even more sophisticated models.
They define 
regularized least squares schemes that 
can outperform conventional 
parametric identification \cite{SurveyKBsysid,LCB2019,BAHP16,mkcdc12,Pillonetto2016}.  
All of these kernels belong to the more general class 
of (BIBO) stable kernels that 
induce RKHSs of absolutely summable impulse responses. 
A fundamental characterization of these kernels has been known in the literature
at least since 2006 \cite{Carmeli}. It says that a kernel $K$ is stable if and only if it induces a bounded integral operator
mapping the space $\ell_{\infty}$ of 
essentially bounded functions into the space $\ell_1$ of absolutely summable functions,
see also \cite{DinuzzoSIAM15,ChenStableRKHS}.
This result is the starting point of this paper.
Building upon it, 
new structural properties of stable RKHSs will be obtained working in 
discrete-time ($\mathcal{X}$ becomes the set of natural numbers).
In particular, the main contributions of the paper 
are the following ones.\\

First, we will obtain fundamental RKHSs inclusion properties that shed new light 
on the relationships between stable kernels and e.g. absolutely summable
and finite-trace kernels. This result defines in a natural and simple way 
new stability tests on several classes of kernels.
It also contains, as immediate corollaries, some instability results obtained in the literature 
through ad-hoc theorems, e.g. regarding the translation invariant class that contains the 
popular Gaussian kernel \cite{Minh2010,DinuzzoSIAM15,PillInsights2018}.\\

As for the second contribution, let $\ell_2$ be the space of squared summable sequences with the usual inner product.
Then, we show that any stable kernel admits a \emph{spectral (Mercer) feature map} 
$\phi: \mathcal{X} \rightarrow \ell_2$ with
$$
\phi(x) = \{ \sqrt{\lambda_i} \rho_i(x)  \}_{i=1}^{\infty} 
$$
where $\lambda_i$ and $\rho_i$ are, respectively, the kernel eigenvalues and eigenfunctions 
forming an orthonormal basis in $\ell_2$.
The fact that any stable RKHS is generated 
by such basis 
provides the fundamental link with the important literature on impulse response estimation via
orthonormal functions 
\cite{WalLag1991,WalIFAC1994,NinnessOrth1999,OrthBook2005}.
Furthermore, under an algorithmic viewpoint, many efficient machine learning procedures exploit truncated Mercer expansions
for approximating the kernel, e.g. see \cite{Williams:2000,Kumar:2012,Gittens:2016} and also \cite{Zhu98,PillPAMI2}  for discussions
on their optimality in a stochastic framework.
These works trace back to the so-called Nystr\"{o}m method 
where an integral equation is replaced by finite-dimensional approximations \cite{Atkinson1975,Baker1977}.
For system identification, the works  \cite{PilAuto2007,CarliIFAC12} have shown that
a relatively small number of eigenfunctions (w.r.t. the data set size) can capture impulse responses regularized estimates.
Our result 
shows that any stable kernel is amenable to these fast computational schemes.
To exploit them, closed-form expressions
of the  $\lambda_i$ and $\rho_i$ are desirable. Determining the
spectrum of $K$ is however far from trivial in general but
numerical approximations can be adopted.
In this regard,  we show that singular value decompositions applied to
truncated stable kernels 
generate a sequence of spectra convergent in $\ell_2$ to the correct one.
This can be seen as a novel convergence result for a Nystr\"{o}m-type method on unbounded 
kernel domains.


Third, having established that any stable RKHS is generated 
by a basis of $\ell_2$, 
the question is however which 
kind of orthonormal functions and of their combinations lead to stable kernels.
This motivates the study of stability conditions for 
models built through 
feature maps and (\ref{EQ01}).
This route loses the advantage of implicit encoding
since it rarely leads to closed-form kernel expressions
(this is the dual problem of the Mercer expansion).
However, such issue is very relevant. In fact,  
recent 
literature has shown that 
important dynamic system features, like the presence of resonances,
can be conveniently described using feature maps
e.g. induced by Kautz models \cite{ChenOrth2015,DARWISH2018318}. 
Then, we will provide the necessary and sufficient stability condition 
for kernels defined by Mercer expansions.
This new outcome should be taken into account
when formulating any linear system model 
whose aim is to combine orthogonal basis functions in $\ell_2$ with stability information via 
kernel-based regularization.\\

So, overall, our results 
have impact on stability tests, impulse responses modeling and computational efficiency issues.
To illustrate them, the paper is organized as follows.  
Section \ref{Sec2} reports a brief overview on (stable) RKHSs setting up also some notation.
In Section \ref{Sec3} we obtain new inclusion properties of some notable kernels classes that provide 
fundamental insights on the structure of stable kernels.
Section \ref{Sec4} shows that any stable kernel admits a Mercer expansion in $\ell_2$
and discusses the link with impulse response estimation via orthonormal bases.
It also shows how to numerically recover kernel eigenfunctions and eigenvalues.
In Section \ref{Sec5} the necessary and sufficient condition for RKHS stability in the Mercer feature space
is worked out.
Conclusions then end the paper while the proof of all the new theorems are  gathered in Appendix.

\section{Overview on RKHSs and stability condition}\label{Sec2}

We are interested in spaces of functions containing discrete-time
impulse responses of causal systems. 
Hence, the function domain is the set of natural numbers $\mathbb{N}$. 
In addition, the elements of the space can be also seen as sequences containing impulse response coefficients.\\
We will consider in particular the so-called Reproducing Kernel Hilbert Spaces (RKHSs).  They are 
in one-to-one correspondence with positive semidefinite kernels that, in our setting, map $\mathbb{N} \times \mathbb{N}$
into the real line. However, 
in view of the nature of the domain, in what follows it is more convenient to see the kernel as an infinite-dimensional matrix
with the $(i,j)$-entries denoted by $K_{ij}$. The positive semidefinite constraints then imply that, for any choice of 
integers $\{p_1,\ldots,p_m\}$, the $m \times m$ matrix $A$, with  $A_{ij} = K_{p_i p_j}$, is symmetric and positive semidefinite.\\
As already recalled, the RKHS inherits the properties of a kernel. Indeed, the values $K_{ij}$ can be interpreted 
as a similarity measure between the $i$-th and the $j$-th element of the sequence. In linear system identification,
the interest is in particular addressed to (BIBO) stable kernels. 
They induce RKHSs containing
only absolutely summable vectors.
To introduce them, let $\ell_\infty$ and $\ell_1$ be the spaces of
bounded and absolutely summable sequences of real numbers, respectively, i.e. 
$$
\ell_\infty = \Big\{ \{u_i\}_{i \in {\mathbb{N}}}  \ \mbox{s.t.} \   \| u \|_\infty <  \infty \Big\},
$$
and
$$
\ell_1 = \Big\{ \{u_i\}_{i \in {\mathbb{N}}}   \ \mbox{s.t.} \   \| u \|_1 <  \infty \Big\},
$$
with
$$
\| u \|_\infty = \sup_{i \in {\mathbb{N}}}  |u_i| \quad \mbox{and} \quad \| u \|_1 = \sum_{i \in {\mathbb{N}}}  |u_i|. 
$$
Now, note also that the kernel $K$ defines an acausal linear time-varying system, often called
kernel operator in the literature: given an input (sequence) $u$,  
the output at instant $i$ is $\sum_{j=1}^\infty K_{ij} u_j$. 
Then, using notation of ordinary algebra to handle infinite-dimensional objects, the 
output can be indicated by $Ku$ with $u$ an infinite-dimensional (column) vector. 
With this in mind, the following fundamental theorem reports the 
necessary and sufficient condition for RKHS stability.

\begin{theorem}[RKHS stability \cite{Carmeli}]\label{NecSuffStabKer}
Let $\mathcal{H}$ be the RKHS induced by $K$. 
Then, it holds that
\begin{equation} \label{CondNS}
 \mathcal{H} \subset \ell_1 \  \iff   \  Ku \in \ell_1 \ \ \forall u \in \ell_{\infty}. 
\end{equation}
\begin{flushright}
$\blacksquare$
\end{flushright}
\end{theorem}

The stability condition is equivalent to requiring that the kernel operator is a bounded (continuous)
map between $\ell_{\infty}$ and $\ell_1$, see \cite{SiamArxivAKS2019}. 

\begin{remark}\label{remark1}
The third fundamental space used in this paper, 
already mentioned in the introduction, is that containing squared summable sequences, i.e.
$$
\ell_2 = \Big\{ \{u_i\}_{i \in {\mathbb{N}}}   \ \mbox{s.t.} \   \| u \|_2 <  \infty \Big\},
$$
with
$$
 \| u \|^2_2 = \sum_{i \in {\mathbb{N}}} u_i^2. 
$$
In particular, two types of kernel operators induced by $K$ will be encountered during our analysis.
The first one is that described above mapping $\ell_{\infty}$ into $\ell_1$  
while the second one is that mapping $\ell_2$ into $\ell_2$ itself.   
\end{remark}

\section{Inclusion properties of some notable kernels classes}\label{Sec3}

In this section we derive 
new relationships between stable kernels and other fundamental classes.
Let $\mathcal{S}_{s}$ be the set containing all the stable RKHSs.
Then, we also  consider
\begin{itemize}
\item the set $\mathcal{S}_{1}$ containing all the RKHSs induced
by absolutely summable kernels, i.e. satisfying the constraint
\begin{equation*}
\sum_{ij} \ |K_{ij}| < +\infty;
\end{equation*}
\item the set $\mathcal{S}_{ft}$ of RKHSs associated to 
finite-trace kernels that are characterized by
\begin{equation*}
\sum_{i} \ K_{ii} < +\infty;
\end{equation*}
\item  the set $\mathcal{S}_{2}$ induced by squared summable kernels, i.e. satisfying
 \begin{equation*}
\sum_{ij} \ K_{ij}^2 < +\infty.
\end{equation*}
\end{itemize}

The following result 
then holds.
\begin{theorem}\label{ThRKHSinclusions}
One has
\begin{equation}\label{RKHSinclusions}
\mathcal{S}_{1} \subset \mathcal{S}_{s} \subset \mathcal{S}_{ft} \subset \mathcal{S}_{2}
\end{equation}
\begin{flushright}
$\blacksquare$
\end{flushright}
\end{theorem}
Fig. 1 provides a graphical description of Theorem \ref{ThRKHSinclusions} in terms of inclusions
of kernels classes. Some comments about its meaning, under the perspective of 
stability tests, are now in order.\\

\begin{figure*}
  \begin{center}
\hspace{.1in}
 { \includegraphics[scale=0.5]{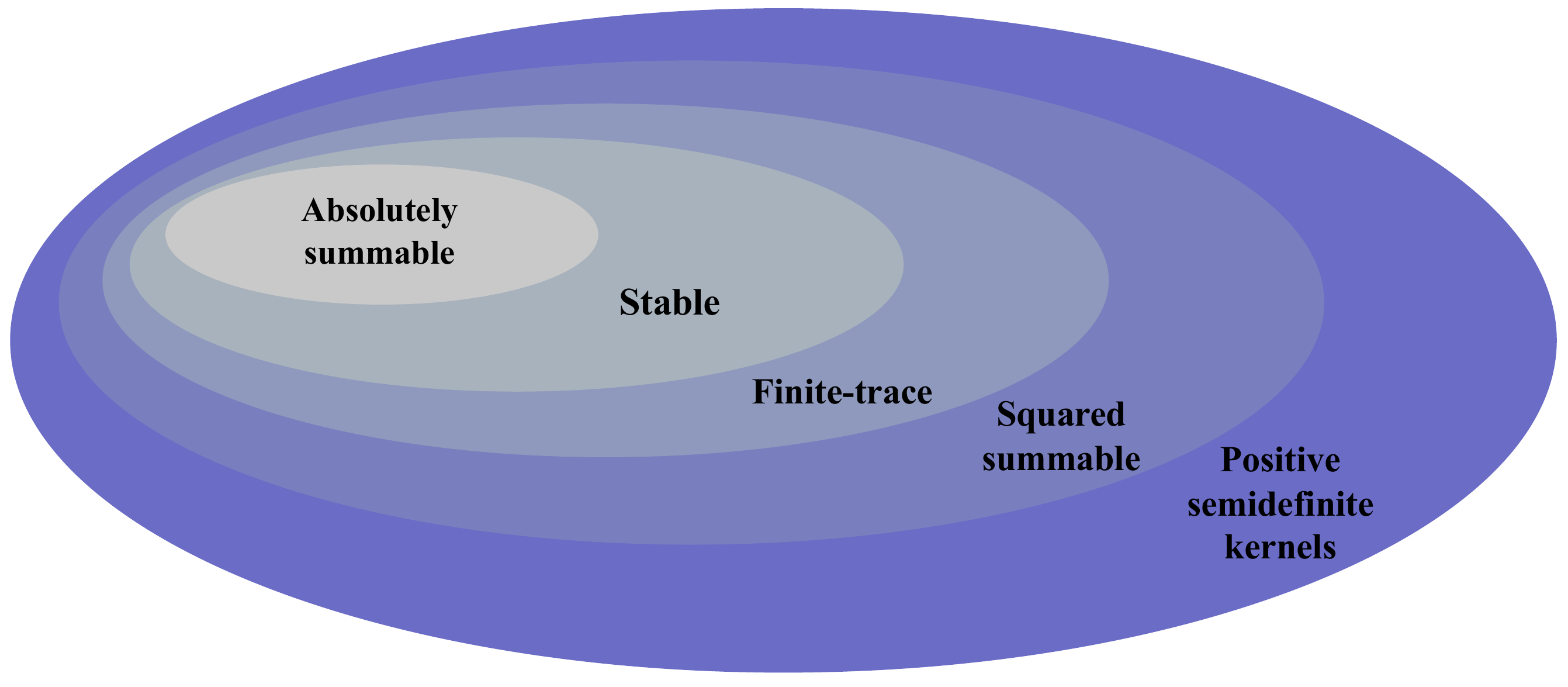}} 
 \caption{Inclusion properties of different kernel classes} 
    \label{Fig1}
     \end{center}
\end{figure*}

Regarding  $\mathcal{S}_{1} \subset \mathcal{S}_{s}$,
while it is trivial to show that kernel absolute summability implies stability, 
the notable fact is that such inclusion is strict.
In this regard, recall that in \cite{Carmeli,DinuzzoSIAM15}, 
immediately after reporting Theorem \ref{NecSuffStabKer}, 
the authors mentioned kernel absolute summability as a sufficient condition, 
with the desire to formulate a (in some sense) simpler stability test. 
Its necessity was however left as an open problem.
Many papers have then cited and exploited kernel summability as a stability check 
without answering such question, e.g. see \cite{Darwish2015,ChenKS2015,Fujimoto2017,ChenKS2018}. 
Theorem \ref{ThRKHSinclusions} points out that the equivalence does not hold.
So, one cannot conclude that a kernel is unstable
from the sole failure of absolute summability.\\

The relation $\mathcal{S}_{s} \subset \mathcal{S}_{ft}$ means that the
set of finite-trace kernels contains the stable class.
Also this inclusion is strict, hence the analysis of the trace is useful 
only to prove that a given RKHS is not contained in $\ell_1$.
This has however interesting consequences. 
For instance, in \cite{Minh2010} the instability of the Gaussian kernel
$$
K_{ij} =  e^{-(i-j)^2} 
$$
was proved exploiting a complex RKHS representation through a 
generalization of the Weyl inner product for the homogeneous polynomial space.
In \cite{DinuzzoSIAM15}[Appendix] this result was greatly extended by proving that
all the RKHSs induced by translation invariant kernels, i.e. of the form 
$$
K_{ij}=h(i-j)
$$
(with $h$ satisfying the positive semidefinite constraints) are not contained in $\ell_1$.
The Schoenberg representation theorem was used, see p. 3309 of \cite{DinuzzoSIAM15}.
All of these ad-hoc theorems are now trivial corollaries of Theorem \ref{ThRKHSinclusions} since
the trace of a translation invariant kernel is $\sum_i \ K_{ii}=\sum_i \ h(0)$ and it always diverges
unless $h$ is the null function. Even more importantly, 
many other instability results become immediately available. 
One can e.g. claim that all the kernels whose diagonal elements 
satisfy $K_{ii} \propto i^{-\delta}$  
are unstable if $\delta \leq 1$.\\

Finally, the strict inclusion $\mathcal{S}_{ft} \subset \mathcal{S}_{2}$ 
shows that a stability check relying on kernel squared summability 
does not make much sense. In fact, the finite-trace test is in any case both more powerful and 
simpler to perform. 

\section{Mercer expansions of stable kernels}\label{Sec4} 

\subsection{Mercer feature maps for stable kernels} 

Recalling also Remark \ref{remark1}, we are now interested
in the operator induced by a stable kernel as a map from $\ell_2$ into itself.\\ 
A kernel operator is compact if it maps any bounded sequence $\{v_i\}$ into a sequence
$\{K v_i\}$ from which a convergent subsequence can be extracted \cite{Rudin,Zeidler}. 
Theorem \ref{ThRKHSinclusions} ensures that any 
stable kernel $K$ is finite-trace. Combining this fact 
with Lemma \ref{lemma4} present in Appendix, one obtains the following result.

\begin{theorem} \label{ThStableCompact}
Any operator induced by a stable kernel is self-adjoint, 
positive semidefinite and compact as a map from $\ell_2$ into $\ell_2$ itself.
\begin{flushright}
$\blacksquare$
\end{flushright}
\end{theorem}

The above theorem is important because it allows us to associate
to any stable kernel a Mercer feature map built through
an orthonormal basis of $\ell_2$. 
In particular, the following result is a direct consequence 
of the spectral theorem \cite{Dunford1963} (applied 
to kernels defined over $\mathbb{N} \times \mathbb{N}$)
that holds 
indeed by virtue of Theorem \ref{ThStableCompact}.

\begin{proposition}[Representation of stable kernels]  \label{ThStableCompact2}
Let $K$ be stable. Then, there always 
exists an orthonormal basis of $\ell_2$ composed by eigenvectors $\{\rho_i\}$ of $K$
with corresponding eigenvalues $\{\lambda_i\}$, i.e.
$$
K \rho_i = \lambda_i \rho_i, \ \ i=1,2,\ldots.
$$
In addition, the spectral Mercer feature map
$\phi: \mathbb{N} \rightarrow \ell_2$ with
$$
\phi(x) = \{ \sqrt{\lambda_i} \rho_i(x)  \}_{i=1}^{\infty} 
$$
is always well-defined and each $(x,y)$ entry of $K$ admits the representation 
\begin{equation}\label{Kexpansion}
K_{xy} = \langle \phi(x), \phi(y) \rangle_2 = \sum_{i=1}^{+\infty} \lambda_i  \rho_i(x) \rho_i(y), 
\end{equation}
where $x,y \in \mathbb{N}$.
\begin{flushright}
$\blacksquare$
\end{flushright}
\end{proposition}

The pointwise convergence of the kernel expansion (\ref{Kexpansion})
stated in Proposition \ref{ThStableCompact2}, combined with the same arguments used in 
\cite{Cucker01}[Chapter 3, Theorem 4] or  \cite{Sun05}[Theorem 1],
allows us to obtain the following characterization of any stable RKHS.

\begin{proposition}[Representation of stable RKHSs] \label{ThStable3}
Let $K$ be stable and assume that any kernel eigenvalue
satisfies $\lambda_i>0$. Then, the stable RKHS associated to $K$
always admits the representation
\begin{equation}\label{Hexpansion}
\mathcal{H} = \Big\{ f = \sum_{i=1}^{\infty} a_i \rho_i  \ \ \text{s.t.} \ \  \sum_{i=1}^{\infty} \ \frac{a^2_i}{\lambda_i} < +\infty \Big\} ,
\end{equation}
where the $\rho_i$ are the eigenvectors of $K$ forming an orthonormal basis of $\ell_2$.
\begin{flushright}
$\blacksquare$
\end{flushright}
\end{proposition}

\begin{remark} In Proposition \ref{ThStable3} we have assumed that all the 
eigenvalues of the stable kernel $K$ are strictly positive so that $\mathcal{H}$ is infinite-dimensional.
If some eigenvalue is null, $\mathcal{H}$ is spanned only by the eigenvectors associated to 
non-null $\lambda_i$. If only a finite number of $\lambda_i$ is different from zero, $K$ is finite-rank and $\mathcal{H}$ is finite-dimensional.
A notable case is that of the RKHSs induced by truncated kernels, i.e. such that there exists $d$ such that $K_{ii}=0 \ \forall i>d$.
This kind of kernels induce finite-dimensional RKHSs containing FIR systems of order $d$. 
\end{remark}

\subsection{Connection with impulse response estimation using orthonormal bases of $\ell_2$} 

As mentioned in Introduction, important impulse response models
exploit orthonormal functions $\{\rho_i\}$ in $\ell_2$ given e.g. by Laguerre or Kautz models \cite{NinnessOrth1999}.
Then, linear least squares  estimators are often adopted to recover the expansion coefficients $\{a_i\}$.
Specifically, let $L_k[f]$ be the system output, i.e. the convolution between the known input and $f$, at 
the instant $t_k$ where the noisy measurement $y_k$ is available.
Then, the impulse response estimate from a data set of size $N$ is 
\begin{subequations} \label{LS}
\begin{align}
\hat{f} &= \sum_{i=1}^d \ \hat{a}_i \rho_i \\
\{\hat{a}_i\}_{i=1}^d &= \argmin_{\{a_i\}_{i=1}^d} \ \sum_{k=1}^N \ \left(y_k - L_k\left[ \sum_{i=1}^d \ a_i \rho_i\right] \right)^2 
\end{align}
\end{subequations} 
where $d$ determines model complexity and is typically selected using AIC or cross validation (CV) \cite{Ljung:99}.\\
An alternative option originally proposed in \cite{SS2010} consists of searching for the impulse response estimate in
a stable and infinite-dimensional RKHS with ill-posedness faced by regularization. The least squares estimator (\ref{LS}) 
is replaced by the following regularized least squares (ReLS) problem
\begin{equation} \label{ReLS}
\hat{f} = \argmin_{f \in \mathcal{H}} \ \sum_{k=1}^N \ \left(y_k - L_k\left[ f \right] \right)^2 + \gamma \| f \|^2_{\mathcal{H}}
\end{equation}
where $\| \cdot \|_{\mathcal{H}}$ is the RKHS norm and the positive scalar $\gamma$ is the so-called 
regularization parameter. It can e.g. be estimated using empirical Bayes approaches, e.g. see \cite{PillonettoMLrob2015}.

The results obtained in the previous subsection permit 
to understand analogies and differences between (\ref{LS})  and (\ref{ReLS}).
In fact, by using Proposition \ref{ThStable3} and recalling from \cite{Cucker01} also that
$$
f = \sum_{i=1}^{\infty} \ a_i \rho_i  \  \implies  \   \| f \|^2_{\mathcal{H}}= \sum_{i=1}^{\infty} \ \frac{a^2_i}{\lambda_i}, 
$$
the following result holds.

\begin{proposition}[Representation of ReLS in stable RKHSs] \label{ThStableReLS}
Let $K$ be stable. Assume that any kernel eigenvalue
satisfies $\lambda_i>0$ and consider the representation (\ref{Hexpansion}) 
of the induced RKHS. Then,
(\ref{ReLS}) is equivalent to
\begin{subequations} \label{ReLS2}
\begin{align}
\hat{f} &= \sum_{i=1}^{\infty} \ \hat{a}_i \rho_i \\
\{\hat{a}_i\}_{i=1}^\infty &= \argmin_{\{a_i\}_{i=1}^\infty} \ \sum_{k=1}^N \ \left(y_k - L_k\left[ \sum_{i=1}^\infty \ a_i \rho_i\right] \right)^2 + \gamma \sum_{i=1}^{\infty} \ \frac{a^2_i}{\lambda_i}.
\end{align}
\end{subequations} 
\begin{flushright}
$\blacksquare$
\end{flushright}
\end{proposition}

So, regularized least squares in a stable (infinite-dimensional) RKHS always model impulse responses using an $\ell_2$ 
orthonormal basis, as in the classical works \cite{WalLag1991,OrthBook2005}.
But the key difference between (\ref{LS})  and (\ref{ReLS2}) is that complexity is no more controlled by the model order
since $d$ is set to $\infty$. It instead depends on the regularization parameter $\gamma$ 
that trades-off data fit and the penalty term.
This latter induces stability
by constraining the decay rate of the expansion coefficients to zero through the kernel eigenvalues $\lambda_i$.\\

The estimator (\ref{ReLS}) would seem a computationally unfeasible
(infinite-dimensional) variational problem.
Actually, according to the representer theorem \cite{Kimeldorf70,Scholkopf01,ArgyriouD2014}, the estimate 
$\hat{f}$ belongs to a subspace of dimension equal to the data-set size $N$. For dynamic systems, 
it is determined by the kernel and the system input, see \cite{SurveyKBsysid}[Part III] for details.
The kernel implicit encoding so pemits to compute the impulse response estimate without
knowing the basis functions $\{\rho_i\}$.
However, achieving the $N$ expansion coefficients requires $O(N^3)$ operations,
so that for large data sets alternative procedures are desirable.
A strategy for approximating  (\ref{ReLS}) is to 
use the equivalence with (\ref{ReLS2}) then
resorting to 
truncated Mercer expansions. 
Specifically, Problem (\ref{ReLS2}) is replaced by the following 
$d$-dimensional surrogate
\begin{subequations} \label{ReLS3}
\begin{align}
\hat{f}^{(d)} &= \sum_{i=1}^{d} \ \hat{a}_i^{(d)} \rho_i \\
\{\hat{a}_i^{(d)}\}_{i=1}^d &= \argmin_{\{a_i\}_{i=1}^d} \ \sum_{k=1}^N \ \left(y_k - L_k\left[ \sum_{i=1}^d \ a_i \rho_i\right] \right)^2 + \gamma \sum_{i=1}^{d} \ \frac{a^2_i}{\lambda_i}.
\end{align}
\end{subequations} 
Note that $d$ has not to trade-off bias and variance here as it instead happens in (\ref{LS}).
It has instead to be sufficiently large so that $\hat{f}^{(d)}$ is close to $\hat{f}$.
Indeed, in \cite{PilAuto2007} it has been shown that convergence holds in the RKHS norm as $d$ grows to infinity.
In addition, in \cite{PilAuto2007,CarliIFAC12} numerical experiments have shown that
a relatively small number of eigenfunctions (w.r.t. the data set size $N$)   
can provide really good approximations. This is advantageous since, after numerically computing
each value $L_k[\rho_i]$, the estimate  $\hat{f}^{(d)}$ requires only $O(Nd^2)$ operations.

\subsection{Numerical recovery of the Mercer $\ell_2$ feature map}

In the previous subsection, we have outlined that
Mercer expansions of $K$
can be important also for implementing ReLS. 
However, obtaining closed form expressions of the Mercer feature map is often prohibitive.  
The following result 
fills this gap
by showing that the $\ell_2$ basis of a stable RKHS and the kernel eigenvalues 
can be numerically estimated (with arbitrary precision) by a sequence of SVDs applied to truncated kernels.
This result is not trivial since, in the literature, 
the problem could not even be posed on a firm theoretical ground. 
In fact, it was not known whether a stable kernel admitted a Mercer expansion in $\ell_2$,
a fact now established by Proposition \ref{ThStableCompact2}.\\ 
Given a kernel $K$, the notation $K^{(d)}$ indicates its truncated version, i.e.
the $d \times d$ matrix obtained by retaining only its first $d$ rows and columns.
Then, $\rho_i^{(d)}$ and $\lambda_i^{(d)}$ are  the eigenvectors (thought of as elements of $\ell_2$ with a tail of zeros) and the 
eigenvalues obtained by the SVD of $K^{(d)}$. Single multiplicity is assumed for each  $\lambda_i$,
see Remark \ref{RemarkEigMult} in Appendix for further discussions.

 \begin{theorem}[Estimation of Mercer expansions in $\ell_2$] \label{SVDconv}
 Let $K$ be stable or, more generally, be a kernel inducing a compact operator as a map from $\ell_2$ into $\ell_2$ itself.
Let also $\rho_i$ and $\lambda_i$ denote, respectively,  its eigenfunctions (forming an orthonormal basis in $\ell_2$) 
and the corresponding eigenvalues. 
 Assume also that the multiplicity of each $\lambda_i$ is equal to one.
 Then, for any $i$,  as $d$ grows to $\infty$ it holds that 
 \begin{subequations} \label{SVDconvEq}
\begin{align}
& \lambda_i^{(d)}  \rightarrow \lambda_i \\
& \| \rho_i^{(d)} - \rho_i \|_2  \rightarrow 0.
\end{align}
\end{subequations} 
where $\| \cdot \|_2 $ is the $\ell_2$-norm.
\begin{flushright}
$\blacksquare$
\end{flushright}
\end{theorem} 

\begin{figure*}
  \begin{center}
   \begin{tabular}{cc} 
 { \includegraphics[scale=0.35]{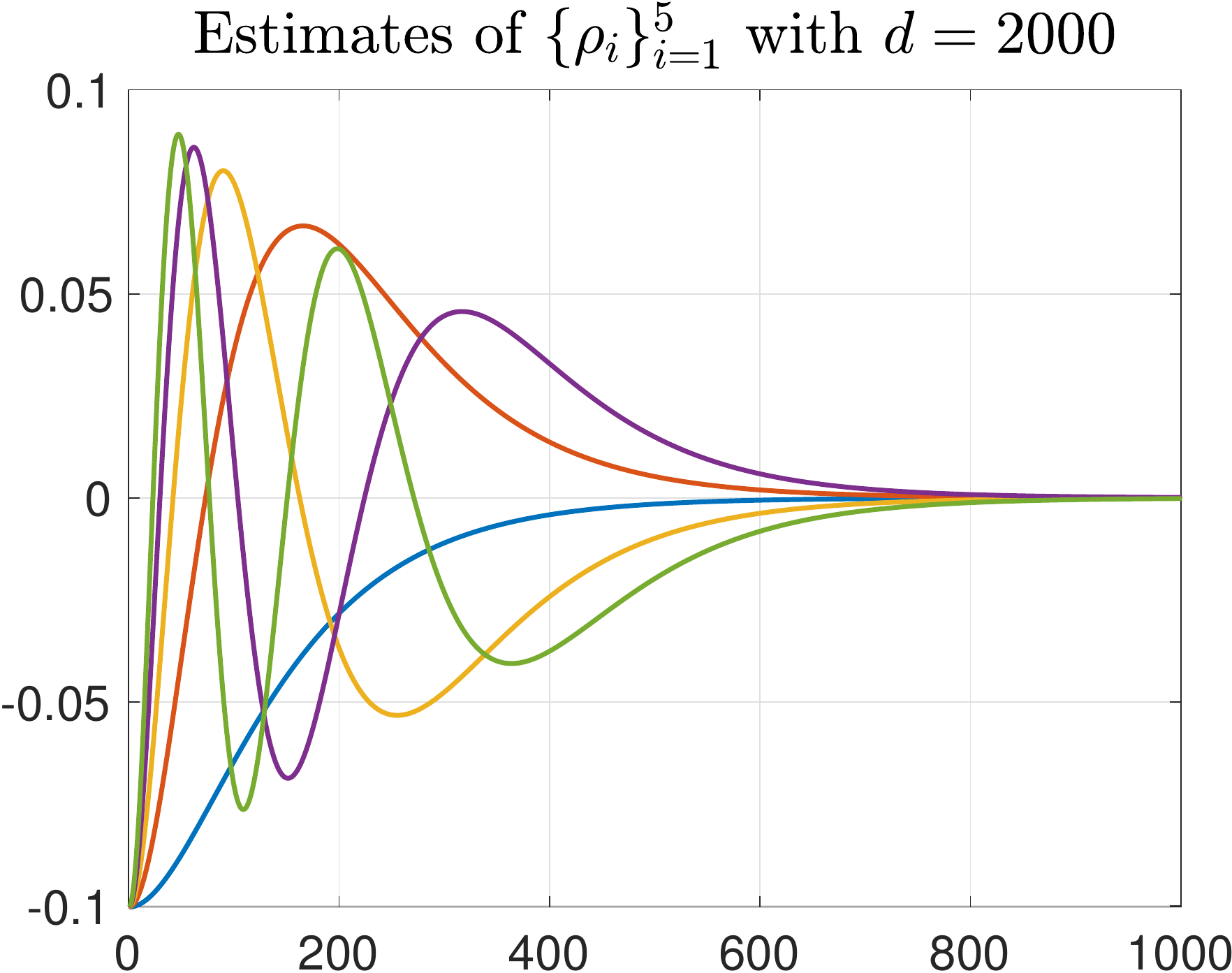}} \  { \includegraphics[scale=0.35]{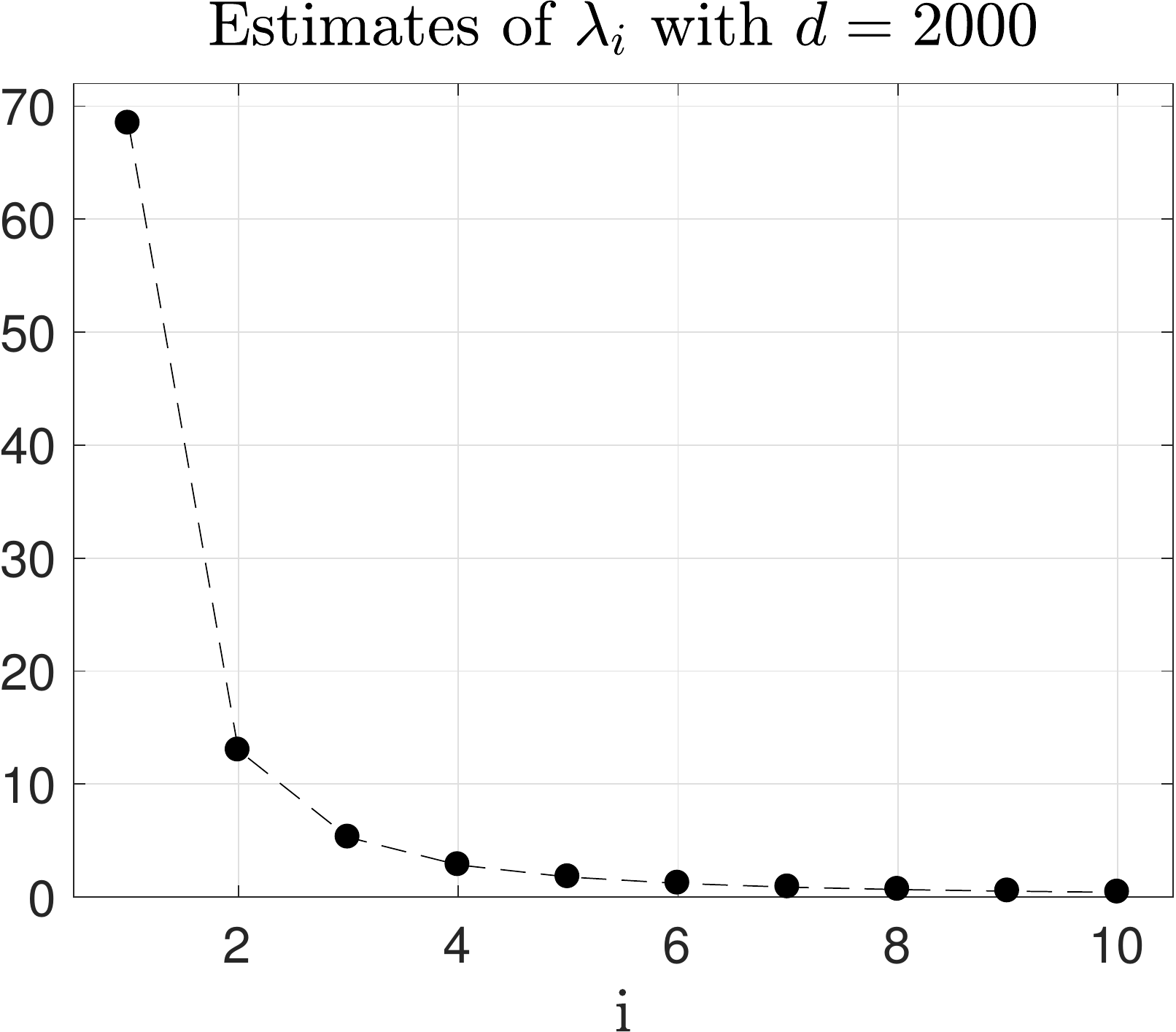}} 
    \end{tabular}
    \caption{Estimates of the first 5 eigenfunctions (left)
and of the first 10 eigenvalues (right) achieved by the SVD of the truncated stable spline kernel with $d=2000$.} \label{Fig2a}
     \end{center}
\end{figure*}
\begin{figure*}
  \begin{center}
   \begin{tabular}{cc} 
 { \includegraphics[scale=0.35]{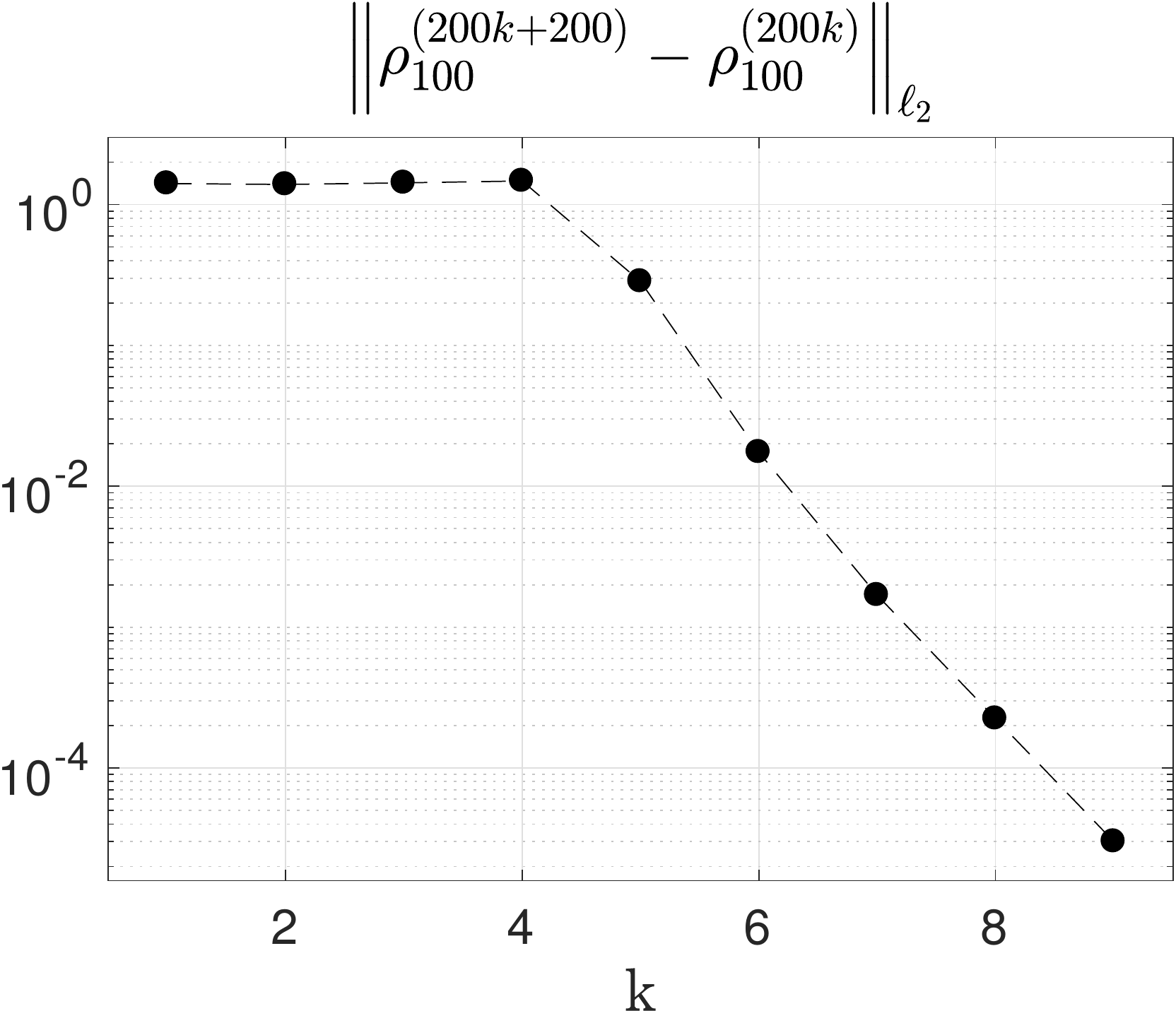}} \  { \includegraphics[scale=0.35]{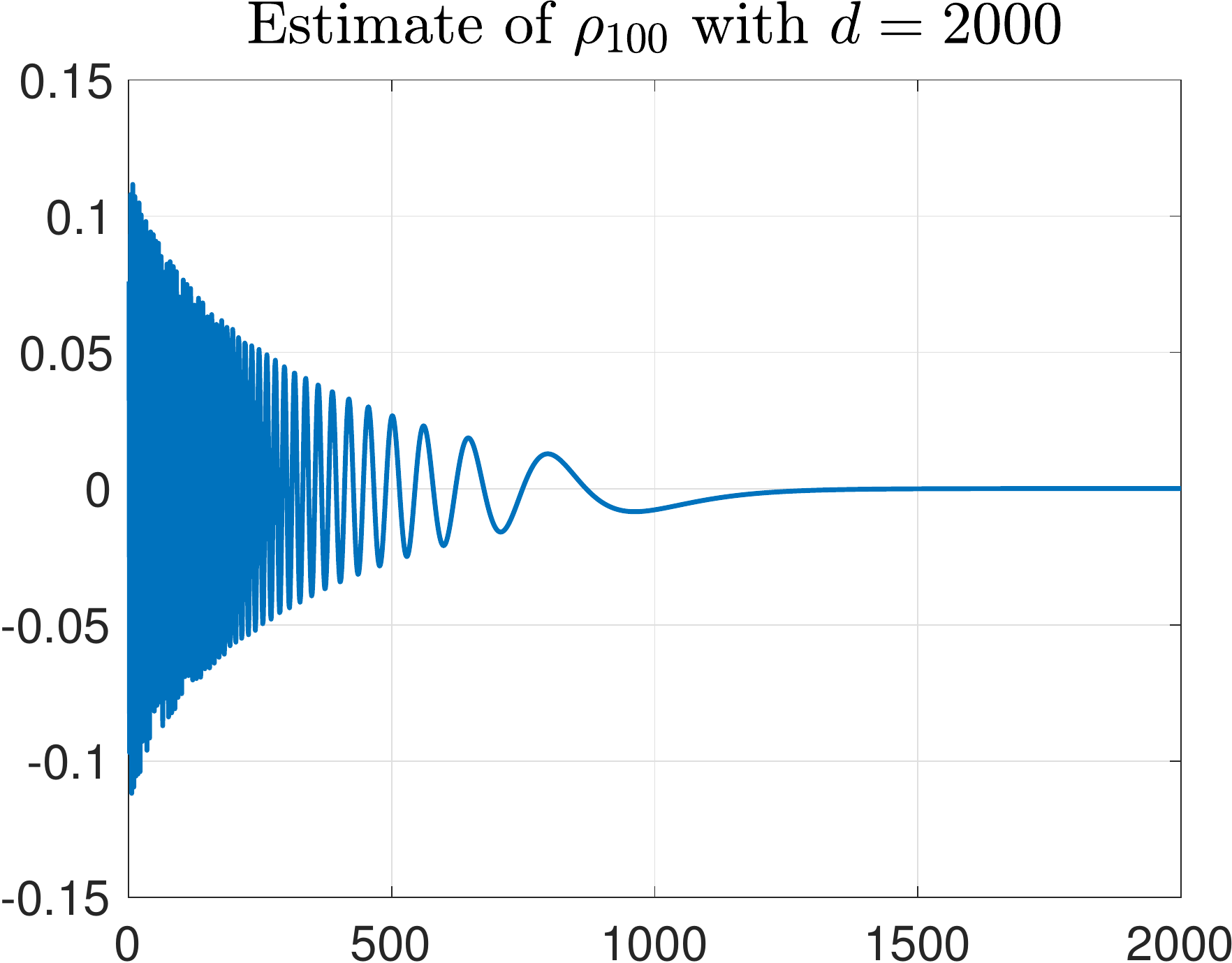}} 
    \end{tabular}
    \caption{Distances measured by the $\ell_2$ norm between $\rho_{100}^{(200k+200)}$ and $\rho_{100}^{(200k)}$ for different $k$ values (left)
    and estimate of $\rho_{100}$ obtained with $d=2000$ (right).} \label{Fig2b}
     \end{center}
\end{figure*}

Theorem \ref{SVDconv} is now applied to the first-order stable spline (SS)
kernel \cite{SS2010}, also called TC kernel in \cite{ChenOL12}. This model is often used to describe
smooth and exponentially decaying impulse responses. Its $(i,j)$ entry is
\begin{equation}\label{SS}
K_{ij} = \alpha^{\max(i,j)}
\end{equation}
where the scalar $0 \leq \alpha <1$ regulates the decay rate of the functions
contained in the induced RKHS. In what follows, we set $\alpha=0.95$.\\
Our aim is to obtain a good approximation of the SS Mercer expansion in $\ell_2$. 
For this purpose, we exploit Theorem \ref{SVDconv} by computing SVDs of truncated SS kernels
of size $d=200,400,\ldots,2000$. Both  $\lambda_i$ and $\lambda_i^{(d)}$
are ordered in decreasing order in what follows.\\
Fig. \ref{Fig2a} plots the estimates of the first 5 eigenfunctions (left)
and of the first 10 eigenvalues (right) achieved with $d=2000$.
The capability of the finite-dimensional
estimator (\ref{ReLS3}) to approximate (\ref{ReLS}) for small values of $d$ will depend on the 
number of data available, the system input and the value of the regularization parameter
$\gamma$. However, the fact that the eigenvalues profile shows that most of the energy of the TC kernel
is captured by the first 5-10 eigenfunctions suggests that values of $d$ much smaller than $N$ can do a good job,
confirming the experimental results reported in \cite{CarliIFAC12}.\\
Fig. \ref{Fig2b} provides some details on the reconstruction of the 100-th eigenfunction as $d$ increases from $200$ to $2000$.
The left panel shows the following $\ell_2$ norms 
$$
\Big \| \rho_{100}^{(200k+200)} - \rho_{100}^{(200k)}   \Big\|_2,
$$  
as a function of the integer $k$. Such norms can be monitored to assess the convergence (ensured by Theorem \ref{SVDconv}) of the 
$\rho_{100}^{(d)}$ 
towards the eigenfunction $\rho_{100}$ of $K$. One can see that for values of $k>4$ the 
discrepancy quickly goes to zero.
The right panel finally plots the approximation of $\rho_{100}$ provided by $\rho_{100}^{(2000)}$.

%
%
%

\section{The necessary and sufficient condition for RKHS stability in the Mercer feature space}\label{Sec5}

So far, our starting point has been a kernel designed
by specifying its entries $K_{ij}$. This modeling approach translates the expected features of
an impulse response into kernel properties,
e.g. smooth exponential decay as described by
(\ref{SS}).
This way takes advantage of basis functions implicit encoding.
Recent literature has shown that also models 
built by designing eigenfunctions $\rho_i$ and eigenvalues $\lambda_i$
are valuable. In fact, kernels relying on 
Laguerre or Kautz functions, that belong to the more general class of 
Takenaka-Malmquist orthogonal basis functions \cite{OrthBook2005}, 
are useful to describe oscillatory behavior or presence of fast/slow poles.\\
This fact motivates the following fundamental problem.
Assigned an orthonormal basis $\{\rho_i\}$ of $\ell_2$,
e.g. of the Takenaka-Malmquist type,
which conditions on the eigenvalues $\lambda_i$ ensure that the kernel
$K_{xy} = \sum_{i=1}^{+\infty} \lambda_i  \rho_i(x) \rho_i(y)$ is stable?
One should also expect that, if $\lambda_i>0 \ \forall i$, there 
exist bases that never satisfy such requirement. 
All of these issues find a definite answer in the following result that provides
the necessary and sufficient condition for kernel stability starting from Mercer expansions.

\begin{theorem}[RKHS stability using Mercer feature maps]\label{NecSuffStabKer2}
Let $\mathcal{H}$ be the RKHS induced by $K$ having Mercer expansion
$K_{xy} = \sum_{i=1}^{+\infty} \lambda_i  \rho_i(x) \rho_i(y)$ 
with $\{\rho_i\}$ an orthonormal basis of $\ell_2$. Define also
\begin{equation*}
\mathcal{U}_{\infty} =\Big\{ \ u \in \ell_{\infty}: \ |u(i)|=1, \ \forall i \ge 1 \ \Big\}.
\end{equation*}
Then, it holds that
\begin{equation} \label{CondNS2}
 \mathcal{H} \subset \ell_1 \  \iff   \  \sup_{u \in {\mathcal U}_{\infty}} \sum_i \lambda_i \langle \rho_i, u \rangle_2^2< +\infty
\end{equation}
where $\langle \cdot, \cdot \rangle_2 $ is the inner product in $\ell_2$.
\begin{flushright}
$\blacksquare$
\end{flushright}
\end{theorem}

\begin{figure*}
  \begin{center}
\hspace{.1in}
 { \includegraphics[scale=0.5]{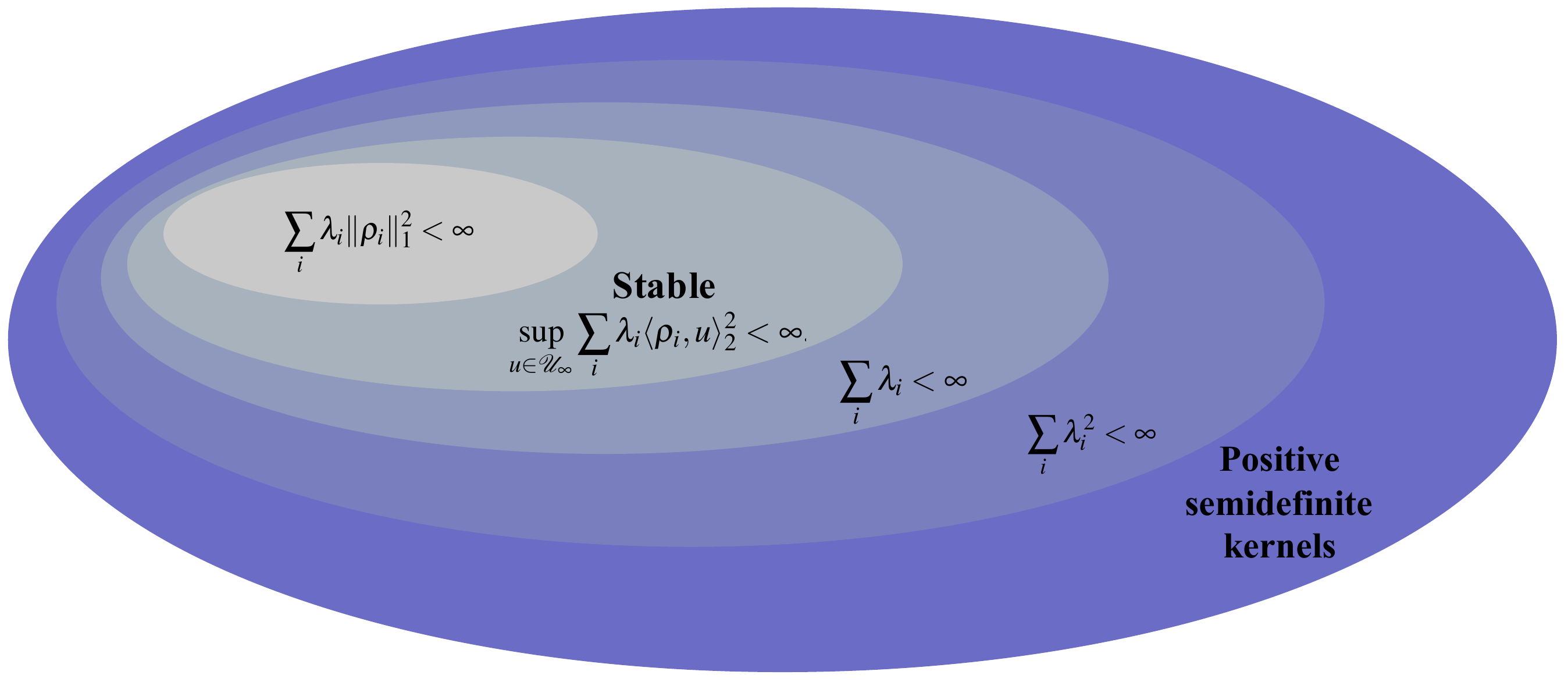}} 
 \caption{Inclusion properties of different kernel classes in terms of the Mercer feature space. This representation is the dual of that reported
 in Fig. \ref{Fig1}. Here the kernel sets are defined through properties of the kernel eigenvectors $\rho_i$, forming an orthonormal basis in $\ell_2$, and of the corresponding kernel eigenvalues $\lambda_i$. The condition $\sum_i \lambda_i \|\rho_i\|_1^2 < \infty$ is the most restrictive since it implies kernel absolute summability. 
The necessary and sufficient condition for stability is $\sup_{u \in {\mathcal U}_{\infty}} \ \sum_i \lambda_i \langle \rho_i, u \rangle_2^2< \infty$. Finally,
$\sum_i \lambda_i < \infty$ and $\sum_i \lambda^2_i < \infty$ are exactly the conditions for a kernel to be finite-trace and squared summable, respectively,
see also the proof of Theorem \ref{ThRKHSinclusions} in Appendix for details.
} 
    \label{Fig1dual}
     \end{center}
\end{figure*}

We discuss some consequences of the above result.\\
If there is one $\rho_i$ corresponding to $\lambda_i > 0$ that doesn't belong to $\ell_1$ then stability is prevented. 
In fact $\langle \rho_i,u \rangle_2= +\infty$ for $u$ containing the signs of the components of $\rho_i$. 
Nothing is however required for the eigenvectors associated to $\lambda_i=0$.\\
Another outcome is the following sufficient stability condition.

\begin{theorem}[Sufficient stability condition using Mercer]\label{NecSuffStabKer3}
Let $\mathcal{H}$ be the RKHS induced by $K$ having Mercer expansion
$K_{xy} = \sum_{i=1}^{+\infty} \lambda_i  \rho_i(x) \rho_i(y)$ 
with $\{\rho_i\}$ an orthonormal basis of $\ell_2$. One has
\begin{equation} \label{CondS2}
\mathcal{H} \subset \ell_1  \impliedby \sum_i \lambda_i \|\rho_i\|_1^2 < +\infty. 
\end{equation}
Moreover, such condition also implies kernel absolute summability and, hence, it is not necessary
for RKHS stability.
\begin{flushright}
$\blacksquare$
\end{flushright}
\end{theorem}

The stability condition (\ref{CondS2}) can be easily used to design 
model stable impulse responses starting from any kind of basis in $\ell_2$.
For instance, as we have recalled in Introduction, Laguerre or Kautz basis functions 
are often used to define $\{\rho_i\}$ that embed information on the dominant pole of the system and/or presence of resonances.
Once such a basis is fixed,  one thus assumes that
the impulse response has the form
$$
f = \sum_{i=1}^{\infty} \ a_i \rho_i. 
$$
Now, to exploit the regularized estimator (\ref{ReLS2})  to identify the system,
the key point is to understand which kind of constraints on the $a_i$ lead to stable models.
The question is equivalent to understand which decay rate of the $\lambda_i$ 
ensures that the regularizer 
$$
\sum_{i=1}^{\infty} \ \frac{a^2_i}{\lambda_i}
$$
enforces impulse responses' absolute summability. 
As a concrete example, Laguerre and Kautz models
all belong to the more general Takenaka-Malmquist class of basis functions 
known to satisfy the constraint
$$
\| \rho_i \|_1 \leq A i,
$$
with $A$ a constant independent of $i$, e.g. see \cite{OrthBook2005}.
Theorem \ref{NecSuffStabKer3} then allows us to immediately conclude that 
the choice 
$$
\lambda_i \propto i^{\nu}, \quad \nu>2
$$
always enforces stability in the estimation process for all the Takenaka-Malmquist class.
So, any stable impulse response model relying e.g. on Laguerre or Kautz
can now embed such a constraint on the eigenvalues' decay.\\

%
%
%
%
%
%

If the orthonormal basis 
functions corresponding to strictly positive eigenvalues are all contained in a ball of $\ell_1$, the following result holds.

\begin{theorem}[Stability with bases uniformly bounded in $\ell_1$]\label{NecSuffStabKer4}
Let $K$ be a kernel having Mercer expansion
$K_{xy} = \sum_{i=1}^{+\infty} \lambda_i  \rho_i(x) \rho_i(y)$ 
with $\{\rho_i\}$ an orthonormal basis of $\ell_2$ and $\| \rho_i \|_1 \leq A < +\infty$ 
if $\lambda_i>0$ ($A$ is a constant independent of $i$). 
Then, one has
\begin{equation} \label{CondS3}
\mathcal{H} \subset \ell_1  \iff \sum_i \lambda_i < +\infty. 
\end{equation}
\begin{flushright}
$\blacksquare$
\end{flushright}
\end{theorem}

Finally, all the new insights on kernel stability in the Mercer feature space are graphically depicted
in Fig. \ref{Fig1dual}.

\section{Conclusions} 

The results reported in this paper shed new light on the RKHSs containing absolutely summable impulse responses.
The inclusion properties here derived give a clear picture on the relationship 
between stable kernels and other fundamental classes. They have important consequences 
for stability tests. In addition, they provide representations of RKHSs and of related regularized least squares estimators
that clarify the relationship with linear system identification via orthonormal bases in $\ell_2$.
Our analysis includes also the necessary and sufficient stability condition for kernels built through these functions.
The paper thus provides new mathematical foundations of stable RKHSs with impact also on stable impulse responses
modeling and estimation.

\section{Appendix}

In what follows, given a finite- or infinite-dimensional matrix $A$, 
the notation  $A \succeq 0$ will indicate that the matrix is symmetric and positive semidefinite. 
 Moreover, the canonical basis in $\ell_2$ will be denoted by $\{e_i\}, i \in {\mathbb N}$.\\
We will also often use $M_m$ to denote a matrix of size $m \times m$.
In addition, let
\begin{equation}\label{OpNorm}
\|M_m\|_{\infty,1}:=\max_{\|u\|_{\infty}=1} \ \|M_m u\|_1,
\end{equation}
that corresponds to  the norm of the linear operator $M_m:{\mathbb R}^m \rightarrow {\mathbb R}^m$ with
the domain and co-domain equipped, respectively, with the $\ell_{\infty}$ and the $\ell_1$ norms.\\
In this Appendix, we will also adopt a different notation for a kernel and a kernel operator 
(so far indicated indistinctly with $K$). 
The notation $M$ will indicate an infinite-dimensional matrix representing a kernel, i.e.  $M \succeq 0$.
The associated kernel operator is $\mathcal{M}$. This will thus be a self-adjoint positive semidefinite operator
with domain and co-domain specified later on.\\  
In addition, given any integer $r \ge 1$, with possibly also $r=\infty$, the set ${\mathcal U}_r$ is defined as follows
\begin{equation}\label{mathU}
{\mathcal U}_r := \{ \ x \in {\mathbb R}^r: x(i)=\pm 1, \forall \ i=1,\dots,r \ \}.  
\end{equation}

\subsection{Proof of Theorem \ref{ThRKHSinclusions}}

 Let $p$ an integer ($p \ge 1$) that also defines  
the odd number  $m=2p+1$ and the corresponding power of two $n=2^m$. 
Let also $x_i \in {\mathcal U}_m$ ($i=1,2,\dots,n$) that, according to (\ref{mathU}), are distinct vectors containing exactly $m$ elements $\pm 1$
(the ordering of such vectors is irrelevant). Then, for any $n=2^3, 2^5, 2^7, \dots$, 
$V^{(n)}$ indicates the matrix of size $n \times m$ given by
\begin{equation}\label{Vn}
V^{(n)}=\left[\begin{matrix} x_1 & x_2 & \dots & x_n \end{matrix} \right]^{\top}.
\end{equation}
Thus, the rows of such matrices contain all the possible permutations of $\pm 1$.
As an example, setting $p=1$ and, hence, $m=3,n=2^3=8$, one obtains
{\footnotesize{
\begin{equation}\label{VnExample} 
V^{(8)}=\left(\begin{array}{ccc}1 & 1 & 1 \\1 & 1 & -1 \\1 & -1 & 1 \\1 & -1 & -1 \\-1 & 1 & 1 \\-1 & 1 & -1 \\-1 & -1 & 1 \\-1 & -1 & -1\end{array}\right)
\end{equation}
}}



\noindent {\bf 1. $\mathcal{S}_{1} \subset \mathcal{S}_{s}$}

The inclusion $\mathcal{S}_{1} \subseteq \mathcal{S}_{s}$ is immediate and well-known in the literature,
as also discussed after stating Theorem \ref{ThRKHSinclusions}.
The (strict) inclusion $\mathcal{S}_{1} \subset \mathcal{S}_{s}$ is not trivial and its proof 
can be found in \cite{SiamArxivAKS2019}. It relies on the building of a particular kernel, depending on the matrices
$V^{(n)}$ defined in (\ref{Vn}), that is stable but non absolutely summable. 

\noindent {\bf 2. $\mathcal{S}_{s} \subset \mathcal{S}_{ft}$}

\begin{lemma}\label{lemma1} \ Let $M_m=M_m^T \succeq 0$ of size $m \times m$. Then the following inequalities hold true
$$
\text{tr} (M_m) \le \|M_m\|_{\infty,1} \le n \ \mbox{tr}(M_m)
$$
\end{lemma}

\begin{proofthm} \ \ \ Using the same arguments contained in the proof of Lemma 3.1 in \cite{SiamArxivAKS2019}, it holds that
$$
\|M_m\|_{\infty,1}=\max_{x \in {\mathcal U}_m} \ \|M_mx\|_1
$$
Recalling that $V^{(n)\top}$ contains all the vectors in ${\mathcal U}_m$ as columns, the problem corresponds to
evaluating 
$$
M_mV^{(n) \top}
$$
and looking for the column with maximum $\ell_1$ norm. The $\ell_1$ norm of any column is easily obtained by means of a scalar product of the column itself with a suitable $x \in {\mathcal U}_m$ corresponding to the signs of the column entries. So, one has that 
$$
V^{(n)}M_mV^{(n) \top}
$$
surely contains (within its $n^2$ entries) these $n$ $\ell_1$ norms. Also, the searched maximum $\ell_1$ norm coincides with the maximum of all $n^2$ entries since $x_1^Tc \le x_2^Tc, \ \forall x_1 \in {\mathcal U}_m$ if $x_2=\mbox{sign}(c )$ (where, for each entry of $c$, \mbox{sign} returns 1 if it is larger than zero and -1 otherwise). 
Now, $V^{(n)}M_mV^{(n) \top} \succeq 0$, which implies that its maximum entry appears along its diagonal, so
$$
\|M_m\|_{\infty,1}=\max_{i=1,\dots,n} \ [V^{(n)}M_mV^{(n) \top}]_{ii}
$$
Now, note that the trace of $V^{(n)}M_mV^{(n) \top}$ satisfies
$$
\text{tr}[V^{(n)}M_mV^{(n) \top}] \ge \|M_m\|_{\infty,1} \ge \frac{1}{n} \ \text{tr}[V^{(n)}M_mV^{(n) \top}].
$$
Finally, 
\begin{eqnarray*}
\text{tr}[V^{(n)}M_mV^{(n) \top}]&=&\text{tr}[M_mV^{(n) \top}V^{(n)}] \\
&=& \text{tr}[M_m (nI_m)]=n \ \text{tr}[M_m]
\end{eqnarray*}
and this concludes the proof. 
\end{proofthm}
\medskip

\begin{lemma}\label{lemma2} \ Let ${\mathcal M}: \ell_{\infty} \rightarrow \ell_1$ be a self-adjoint, positive semidefinite and bounded operator.
Then ${\mathcal M}$ is finite-trace, too.
\end{lemma}

\begin{proofthm} \ \ \ By denoting with $M_k$ the $k \times k$ submatrix of the infinite matrix 
that represents the operator ${\mathcal M}$, it is easy to see that
$$
\|M_k\|_{\infty,1} \le \|{\mathcal M}\|_{\infty,1}<+\infty, \ \forall k=1,2,\dots,
$$
where $\|{\mathcal M}\|_{\infty,1}$ is the operator norm of ${\mathcal M}$.
So, exploiting Lemma \ref{lemma1}, we also have
$$
\text{tr}[M_k] \le \|{\mathcal M}\|_{\infty,1}, \ \forall k=1,2,\dots
$$
Then, since $\text{tr}[M_k]$ is a monotone non-decreasing sequence 
upper-bounded by $\|{\mathcal M}\|_{\infty,1}<+\infty$, letting $M_k(k,k)$
the $k$-th element along the diagonal of $M_k$,
this implies that
$$
\text{tr}[{\mathcal M}] := \sum_{k=1}^{+\infty} M_k(k,k) \le \|{\mathcal M}\|_{\infty,1}<+\infty.
$$
\end{proofthm}
\medskip

Lemma \ref{lemma2} thus shows that $\mathcal{S}_{s} \subseteq \mathcal{S}_{ft}$.
To prove that the inclusion is strict, it suffices to consider the infinite-dimensional matrix (kernel)
$$
M=vv^T.
$$
In fact, one has $\mbox{tr}(M) = \|v\|_2^2<+\infty$ iff $v \in \ell_2$. If $v \notin \ell_1$,
letting $w=sign(v) \in \ell_{\infty}$ one obtains $Mw=v\|v\|_1=\infty$. This proves that the kernel $M$
is unstable.
\medskip

\noindent {\bf 3. $\mathcal{S}_{ft} \subset \mathcal{S}_{2}$}

%
%
%

Given a kernel, represented by an infinite-dimensional matrix $M=M^{\top} \succeq  0$, 
it is now important to consider the induced kernel operator ${\mathcal M}$ as a map from $\ell_2$ into itself.
Its operator norm is given by 
$$
\|{\mathcal M}\|_2 = \sup_{\|v\|_2=1} \ \|Mv\|_2.
$$
In addition, given any orthonormal basis $\{v_i\}$, its nuclear norm is
\begin{equation}\label{trace1}
\sum_{i \geq 1} \langle v_i,Mv_i \rangle_2, 
\end{equation}
while its (squared) Hilbert-Schmidt (HS) norm is
\begin{equation}\label{HSsqnorm}
\sum_{i=1}^{\infty} \ \| M v_i \|_2^2 
\end{equation}
and both of them turn out independent of the particular chosen basis.\\
This view allows us to cast the relation between $\mathcal{S}_{ft}$ and $\mathcal{S}_{2}$
in terms of the important nuclear 
and HS operators.
First, we briefly recall some fundamental results 
that can be found e.g. in \cite{Lidskii1959,Dunford1963,Robert2017}.
By definition, an operator is HS if (\ref{HSsqnorm}) is finite.
In particular, a kernel operator $\mathcal{M}$ is HS iff 
it is induced by a squared summable kernel. 
An HS kernel operator is always compact with squared norm (\ref{HSsqnorm}) given by
\begin{equation}\label{HSsqnorm2}
\sum_{i \geq 1} \ \lambda^2_i
\end{equation}
where the $\lambda_i$ are the eigenvalues of $M$.\\
An operator is said to be nuclear 
if it can be written as the composition of two HS operators.
Any nuclear operator is HS. Also,
a positive semidefinite operator is nuclear if and only  it is compact with
\begin{equation}\label{trace2}
\sum_{i \geq 1} \ \lambda_i < +\infty,
\end{equation}
where the $\lambda_i$ still denote the eigenvalues of $M$  \cite{Robert2017}[Section 2].\\
Now, we want to prove that all the operators induced by the kernels in $\mathcal{S}_{ft}$
are nuclear. 
For any $n \ge 0$, define the projection operators ${\mathcal P}_n$ and
the partial traces $\mbox{tr}_n({\mathcal M})$ as follows
$$
{\mathcal P}_n:\sum_{h=1}^{+\infty} \ a_he_h \rightarrow \ \sum_{h=1}^n \ a_he_h
$$
$$
\mbox{tr}_n({\mathcal M}):=\sum_{h=n+1}^{+\infty} \ M(h,h).
$$
%
%
%

\begin{lemma} \label{lemma3} Let ${\mathcal P}_nv=0$. Then
$$
\|{\mathcal M}v\|_2 \le \sqrt{\mbox{tr}_n({\mathcal M})}\sqrt{\mbox{tr}({\mathcal M})}\|v\|_2
$$
\end{lemma}
\begin{proofthm} \ \ \ We have $v=\sum_{h=n+1}^{+\infty} \ a_he_h$, so
{\scriptsize{
\begin{eqnarray*}
&& \|{\mathcal M}v\|_2^2=\sum_{r=1}^{+\infty} \ \left|\sum_{s=n+1}^{+\infty} \ M(r,s)a_s\right|^2\\ 
&&\le \sum_{r=1}^{+\infty} \ \left(\sum_{s=n+1}^{+\infty} \ |M(r,s)|\cdot |a_s|\right)^2 
\le \sum_{r=1}^{+\infty} \ \left(\sum_{s=n+1}^{+\infty} \ a_s^2 \cdot \sum_{s=n+1}^{+\infty} \ M^2(r,s)\right)\\
&&=\sum_{s=n+1}^{+\infty} \ a_s^2 \ \left(\sum_{r=1}^{+\infty}\sum_{s=n+1}^{+\infty} \ M^2(r,s)\right)
=\|v\|_2^2 \ 
\left(\sum_{r=1}^{+\infty}\sum_{s=n+1}^{+\infty} \ M^2(r,s)\right) \\
&&\le | \|v\|_2^2 \ \left(\sum_{r=1}^{+\infty}\sum_{s=n+1}^{+\infty} \ M(r,r)M(s,s)\right)
=\|v\|_2^2 \ \sum_{r=1}^{+\infty} \ W(r,r) \ \sum_{s=n+1}^{+\infty} \ W(s,s) \\
&&=\mbox{tr}_n({\mathcal W}) \ \mbox{tr}({\mathcal W}) \ \|v\|_2^2
\end{eqnarray*}
}}
and this completes the proof.
\end{proofthm}
\medskip

\begin{lemma} \label{lemma4}
Any finite-trace kernel operator from $\ell_2$ into itself is compact.
\end{lemma}

\begin{proofthm} \ \ \ Let $(w^{(h)}, \ h \in {\mathbb N})$ be a bounded sequence of $\ell_2$ elements, i.e. $\|w^{(h)}\| \le A$ for some $A>0$ and for any $h \in {\mathbb N}$.
Consider the sequence ${\mathcal P}_nw^{(h)}$ that belongs to a subspace isomorphic to ${\mathbb R}^n$.
So, a subsequence $(w^{(h)}, \ h \in {\mathcal I}_n)$ exists such that $({\mathcal P}_nw^{(h)}, \ h \in {\mathcal I}_n)$ converges to some $v_n=\sum_{h=1}^n \ a_he_h$, with $\|v_n\|_2 \le A$ by $\|{\mathcal P}_n w^{(h)}\|_2 \le \|(w^{(h)}\|_2 \le A$. Now we can proceed inductively as follows.  The $({\mathcal P}_{n+1}w^{(h)}, \ h \in {\mathcal I}_n)$ are equal to $({\mathcal P}_nw^{(h)}, \ h \in {\mathcal I}_n)$, except for the $(n+1)-$entry, which is upper bounded (in absolute value) by $A$. 
So, a subsequence $(w^{(h)}, \ h \in {\mathcal I}_{n+1})$ of the $(w^{(h)}, \ h \in {\mathcal I}_n)$ exists such that $({\mathcal P}_{n+1}w^{(h)}, \ h \in {\mathcal I}_{n+1})$ converges to some $v_{n+1}=\sum_{h=1}^{n+1} \ a_he_h$. Note 
that the first $n$ entries are exactly the same for both $v_{n+1}$ and $v_n$, so another $a_i$'s coefficient has been added without modifying the first $n$ coefficients. In this way, we can finally obtain a vector $v \in \ell_2$
$$
v=\sum_{h=1}^{+\infty} \ a_he_h, \ \|v\|_2 \le A
$$
which is the limit in $\ell_2$ of the sequence $v_n$, with $\|v_n\|_2$ forming a monotone non-decreasing sequence upper bounded by $A$. Hence, also $\|v\|_2 \le A$. We have now
\begin{eqnarray*}
\|{\mathcal M}w^{(h)}-{\mathcal M}v\|_2&=&\|{\mathcal M}({\mathcal I}-{\mathcal P}_n)w^{(h)}\\
&+&[{\mathcal M}{\mathcal P}_nw^{(h)}-{\mathcal M}v_n] + {\mathcal M}(v_n-v)\|_2\\
&\le& \|{\mathcal M}({\mathcal I}-{\mathcal P}_n)w^{(h)}\|_2\\
&+&\|{\mathcal M}({\mathcal P}_nw^{(h)}-v_n)\|_2+\|{\mathcal M}(v_n-v)\|_2
\end{eqnarray*}
where ${\mathcal P}_n({\mathcal I}-{\mathcal P}_n)w^{(h)}=0$, for $h \in {\mathcal I}_n$, so Lemma 3 applies leading to $\|{\mathcal M}({\mathcal I}-{\mathcal P}_n)w^{(h)}\|_2 \le \sqrt{\mbox{tr}_n({\mathcal M})}\sqrt{\mbox{tr}({\mathcal M})}$, for $h \in {\mathcal I}_n$. Therefore the first and the third term are both infinitesimal w.r.t. $n$ for any $h \in {\mathcal I}_n$.
The second term is also infinitesimal w.r.t. $h \in {\mathcal I}_n$ for any $n$ since,
using Lemma \ref{lemma3} with $n=0$, for any $v \in \ell_2$ one has 
$$
\|{\mathcal M}v\|_2 \le \mbox{tr}({\mathcal M}) \ \|v\|_2.
$$
Let now $\epsilon_k$ be any monotone non-increasing sequence converging to zero, 
and let $n(k)$ be such that the first and the third term are both less than $\frac{\epsilon_k}{3}$ for any $h \in {\mathcal I}_{n(k)}$.
Let also $h(k) \in {\mathcal I}_{n(k)}$ be such that the second term is less than $\frac{\epsilon_k}{3}$. Thus, we have
$$
0 \le \|{\mathcal M}w^{(h(k))}-{\mathcal M}v\|_2 < \epsilon_k.
$$
In this inductive procedure w.r.t. $k$, we only need to choose $h(1)<h(2)<\dots<h(k)<\dots$ and this is always possible since $h(k)$ can be chosen in infinitely many ways in view of the convergence property of the second term w.r.t. $h$. Finally, by defining the countable set ${\mathcal I}:=\{ \ h(1),h(2),\dots,h(k),\dots \ \}$, the subsequence $(w^{(h)}, \ h \in {\mathcal I})$ of the original sequence $(w^{(h)}, \ h \in {\mathbb N})$, satisfies 
$$
0 \le \|{\mathcal M}w^{(h(k))}-{\mathcal M}v\|_2 < \epsilon_k, \ \forall k=1,2,\dots
$$
with $h(k)$ strictly monotone increasing. Since $\epsilon_k$ is infinitesimal, 
${\mathcal M}w^{(h(k))}$ converges to ${\mathcal M}v$.
So, any bounded sequence $w^{(h)} \in \ell_2$ admits a subsequence $w^{h(k)}$ such that ${\mathcal M}w^{h(k)}$ is
convergent in $\ell_2$, proving the compactness of the operator \cite{Rudin,Zeidler}.
\end{proofthm}
\medskip

Combination of the Lemma \ref{lemma4} and of the spectral theorem \cite{Dunford1963} ensures that there exists a complete 
orthonormal basis of $M$ given by eigenvectors $\{\rho_i\}$ of $M$ 
with corresponding eigenvalues denoted by $\{\lambda_i\}$.
Using first the $\{\rho_i\}$ and then the canonical basis $\{e_i\}$ of $\ell_2$ to evaluate the nuclear norm (\ref{trace1}), 
if $M \in \mathcal{S}_{ft}$ one obtains
$$
\sum_{i \geq 1} \ \lambda_i = \sum_{i \geq 1} \ M_{ii} < +\infty.
$$
So, (\ref{trace2}) holds true and $\mathcal{M}$ is a nuclear operator.
Then, the set inclusion immediately derives from the fact that any nuclear operator is also HS.
Such inclusion is obviously strict as the simple example $M=\mbox{diag} \{1,1/2,1/3, \dots, 1/k, \dots\}$ shows.

\noindent {\bf 4. $\mathcal{S}_{2} \subset$ \mbox{Kernels set}}\\ 
The kernels set contains all the positive semidefinite infinite matrices.
The inclusion is then obvious and is strict as proved by the example $vv^T$ with all the 
entries of the infinite-dimensional column vector $v$ equal to 1.

\subsection{Proof of Theorem \ref{SVDconv}}

Given the infinite-dimensional matrix $M=M^{\top} \succeq 0$
associated with a compact operator, let $M_d$ contain the first $d$ rows and columns of $M$.
Then, we will consider the following partition 
$$
M=\left[\begin{matrix} M_d & A_d \cr A_d^T & B_d \end{matrix} \right]
$$
with the eigenvalues of $M$ and $M_d$ denoted, respectively, by 
$$
\lambda_1(M) \ge \lambda_2(M) \ge \dots \ \  \text{and}   \   \    \lambda_1(d) \ge \lambda_2(d) \ge \dots.
$$
For the moment, no assumption on eigenvalues multiplicities is used. In addition,
$\langle \cdot, \cdot \rangle_2$ denotes the inner-product in $\ell_2$ for infinite-dimensional vectors
or in the classical Euclidean space for finite-dimensional ones. The same holds for $\| \cdot \|_2$.  

\begin{lemma} \label{lemma1eig} For any $d \ge k \ge 1$ it holds that 
\begin{subequations} \label{Eqlemma1eig}
\begin{align}
& \max_{<v_h,v_k>=\delta_{hk}} \ \sum_{h=1}^k \ v_h^TMv_h = \sum_{h=1}^k \ \lambda_h(M), \\ 
& \max_{<v_h,v_k>=\delta_{hk}} \ \sum_{h=1}^k \ v_h^TM_d v_h = \sum_{h=1}^k \ \lambda_h(d)
\end{align}
\end{subequations}
\end{lemma}
\begin{proofthm} \ \ We will exploit the spectral theorem that holds true both for $M$ and for $M_d$. 
Using an orthonormal basis, either in $\ell_2$ or in ${\mathbb R}^d$, the two equalities are obtained by choosing $v_h$ as (one of) the eigenvectors corresponding to either $\lambda_h(M)$ or $\lambda_h(d)$. Now, it suffices to prove that any other choice of the $v_h$ does not lead to results larger than the sum of the first $k$ eigenvalues. We can just focus on $M$. Assume that
$v_h=\sum_{i=1}^{+\infty} \ a_{hi}\rho_i$, with $h=1,2,\dots,k$, 
are orthonormal vectors, expressed in terms of the orthonormal basis $\rho_i$ (each $\rho_i$ is associated with $\lambda_i(M)$). Let also  $v_{h+1}, v_{h+2},\dots$ be any completion of the set $\{ \ v_h, \ h=1,2,\dots,k \ \}$ up to an orthonormal $\ell_2$ basis. Since $a_{hi}$, with $h=1,2,\dots,k$, are the first $k$ elements of the $i-$th row of a unitary (infinite) matrix $U$ with columns given by the vectors $v_h$, one has $\sum_{h=1}^k \ a_{hi}^2 \le 1\ \forall i \in {\mathbb N}$ so that
{\scriptsize{
\begin{eqnarray*} 
&& \sum_{h=1}^k \ v_h^TMv_h=\sum_{i=1}^{+\infty} \ \lambda_i(\sum_{h=1}^k \ a_{hi}^2) \le \sum_{i=1}^{k-1} \ \lambda_i(\sum_{h=1}^k \ a_{hi}^2)+\lambda_k(\sum_{i=k}^{+\infty} \ \sum_{h=1}^k \ a_{hi}^2)\\
&&\sum_{i=1}^{k-1} \ \lambda_i(\sum_{h=1}^k \ a_{hi}^2)+\lambda_k(\sum_{h=1}^k \ \sum_{i=k}^{+\infty} \ a_{hi}^2) 
=\sum_{i=1}^{k-1} \ \lambda_i(\sum_{h=1}^k \ a_{hi}^2)+k\lambda_k -\sum_{i=1}^{k-1} \ \lambda_k (\sum_{h=1}^k \ a_{hi}^2)\\
&&=\sum_{i=1}^{k-1} \ (\lambda_i-\lambda_k)(\sum_{h=1}^k \ a_{hi}^2)+k\lambda_k \le \sum_{i=1}^{k-1} \ (\lambda_i-\lambda_k)+k\lambda_k=\sum_{i=1}^k \ \lambda_i
\end{eqnarray*}
}}
which completes the proof. 
\end{proofthm}

\begin{lemma} \label{lemma2eig}  One has  
 \begin{equation}\label{Eqlemma2eig}  
\|A_d w\|_2^2 \le \lambda_1(d) \lambda_M(B_d) \|w\|_2^2
\end{equation}
where $\lambda_M(B_d)$ is the maximum eigenvalue of $B_d$.
\end{lemma}
\begin{proofthm} \ \  From
 $$
\left[\begin{matrix}v^T & w^T\end{matrix}\right]\left[\begin{matrix}M_d & A_d \cr A_d^T & B_d\end{matrix}\right]\left[\begin{matrix}v \cr w\end{matrix}\right] \ge 0 
$$
it follows that $|v^TA_dw|^2 \le (v^T M_dv)(w^TB_dw)$.
Now, by choosing 
$v=\frac{A_d w}{\|A_d w\|_2}$,
the previous inequality becomes $\|A_dw\|_2^2 \le \lambda_1(d) \lambda_M(B_d) \|w\|_2^2$
where we have applied (\ref{Eqlemma1eig}) with $k=1$ to $M_d$ and $B_d$. 
\end{proofthm}

\begin{lemma}\label{lemma3eig}  One has
\begin{equation}\label{Eqlemma3eig}  
\lim_{d \rightarrow \infty} \ \lambda_M(B_d) = 0 
\end{equation}
\end{lemma}
\begin{proofthm} \ 
\ Let $z_d$ be a unit norm eigenvector of $B_d$ corresponding to $\lambda_M(B_d)$, and define $q_d=\left[\begin{matrix}0_d & z_d^T\end{matrix}\right]^T$
where $0_d$ is a column vector with $d$ zero entries. We have $\|q_d\|_2=1$ by construction, and $Mq_d=\left[\begin{matrix}A_dz_d \cr B_dz_d\end{matrix}\right]=\left[\begin{matrix}A_dz_d \cr \lambda_M(B_d)z_d\end{matrix}\right]$ with $\|Mq_d\|_2 \ge \lambda_M(B_d)$. By compactness, there exists $d(k)$ s.t. $Mq_{d(k)} \rightarrow q \in \ell_2$, so that 
$\| \lambda_M(B_{d(k)})q_{d(k)} - ({\mathcal I}-{\mathcal P}_{d(k)})q\|_2$ is converging to zero. In terms of squared norms, this implies that 
 $\lambda^2_M(B_{d(k)}) - \sum_{h=d(k)+1}^{+\infty} \ q^2(h)$ goes to zero but $\sum_{h=d(k)+1}^{+\infty} \ q^2(h)$ is also infinitesimal proving that $\lambda_M(B_{d(k)}) \rightarrow 0$. Since $\lambda_M(B_d)$ is a monotone non-increasing sequence (a fact that can be proved using (\ref{Eqlemma1eig}) with $k=1$ and $M$ replaced by $B_d$), $\lambda_M(B_d)$ is infinitesimal too. 
\end{proofthm}

\begin{lemma}\label{lemma4eig} \ For any $k \ge 1$, one has 
\begin{equation} \label{Eqlemma4eig}
\lim_{d \rightarrow +\infty} \ \max_{\langle s_h,s_k \rangle_2 =\delta_{hk}} \ \sum_{h=1}^k \ s_h^TM_d s_h = \max_{\langle s_h,s_k\rangle_2=\delta_{hk}} \ \sum_{h=1}^k \ s_h^TMs_h.
\end{equation}
\end{lemma}
\begin{proofthm} \ \ As shown in the proof of Lemma \ref{lemma1eig}, the maximum on the r.h.s. exists. Let it be attained for some (orthonormal) vectors $\left[\begin{matrix}v_h^T & z_h^T\end{matrix}\right]^T$ where $v_h$ has dimension $d$ and  $h=1,2,\dots,k$. One thus has
\begin{eqnarray*}
&& \max_{\langle s_h,s_k \rangle_2 =\delta_{hk}} \ \sum_{h=1}^k \ s_h^TMs_h=\sum_{h=1}^k \ \left[\begin{matrix}v_h^T & z_h^T\end{matrix}\right]\left[\begin{matrix}M_d & A_d \cr A_d^T & B_d\end{matrix}\right]\left[\begin{matrix}v_h \cr z_h\end{matrix}\right] \\
&& \qquad = \sum_{h=1}^k \ v_h^T M_d v_h+2\sum_{h=1}^k \ v_h^T A_d z_h+\sum_{h=1}^k \ z_h^T B_d z_h.
\end{eqnarray*}
As $d$ grows to $\infty$, $\|z_h\|_2$ tends to zero while $\|v_h\|_2$ tends to 1. In addition, 
since  $z_h^T B_dz_h \le \lambda_M(B_d)\|z_h\|_2^2$,
from (\ref{Eqlemma2eig}), (\ref{Eqlemma3eig}) and the inequality $z_h^T B_dz_h \le \lambda_M(B_d)\|z_h\|_2^2$ it comes 
out that $v_h^TA_d z_h$ and $z_h^T B_dz_h$ also go to zero.
From the orthonormality constraint  $\langle v_i,v_j \rangle_2=\delta_{ij}-\langle u_i,u_j \rangle_2$, one then has that the vectors $v_i$ tend to become mutually orthonormal. In particular, by applying the Gram-Schmidt orthonormalization procedure to the $v_i$, one obtains $v_i=w_i+\epsilon_i$, with the $w_i$ mutually orthonormal and the $\|\epsilon_i\|_2$ tending to zero. So, the term $v_h^T M_d v_h$ can be written as $w_h^T M_d w_h+\delta_h$, with $\delta_h$ converging to zero. 
Overall, one has 
\begin{eqnarray*}
\max_{<s_h,s_k>=\delta_{hk}} \ \sum_{h=1}^k \ s_h^TMs_h&=&\lim_{d \rightarrow +\infty} \ \sum_{h=1}^k \ w_h^T M_d w_h\\
&\le& \lim_{d \rightarrow +\infty} \ \max_{<s_h,s_k>=\delta_{hk}} \ \sum_{h=1}^k \ s_h^TM_d s_h
\end{eqnarray*}
Now, $\max_{<s_h,s_k>=\delta_{hk}} \ \sum_{h=1}^k \ s_h^TM_d s_h$ is a monotone non-decreasing sequence since maximizing $M_{d+1}$ w.r.t. vectors with
the last entry equal to zero corresponds to maximizing $M_d$. Hence, 
this sequence of maximum values is upper bounded by $\max_{<s_h,s_k>=\delta_{hk}} \ \sum_{h=1}^k \ s_h^TMs_h$. Therefore, we also obtain
$$
\max_{<s_h,s_k>=\delta_{hk}} \ \sum_{h=1}^k \ s_h^TM_d s_h \le \max_{<s_h,s_k>=\delta_{hk}} \ \sum_{h=1}^k \ s_h^TMs_h
$$
which, together with the previous inequality, shows that
$$
\lim_{d \rightarrow +\infty} \ \max_{<s_h,s_k>=\delta_{hk}} \ \sum_{h=1}^k \ s_h^TM_d s_h=\max_{<s_h,s_k>=\delta_{hk}} \ \sum_{h=1}^k \ s_h^TMs_h
$$
\end{proofthm}
Combining (\ref{Eqlemma1eig}) and (\ref{Eqlemma4eig}), one obtains
$$
\lim_{d \rightarrow +\infty} \ \sum_{h=1}^k \ \lambda_h(d) = \sum_{h=1}^k \ \lambda_h(M)
$$
with convergence in a monotone non-decreasing sense. 
Such relation, evaluated for $k=1$, implies that $\lambda_1(d)$ tends to $\lambda_1(M)$, and then, for $k=2$, $\lambda_2(d)$ tends to $\lambda_2(M)$, and so on, inductively.\\
Let's now consider a unit norm eigenvector corresponding to $\lambda_h(M)$ composed by the subvectors $v_d$ and $w_d$, i.e.
\begin{equation}\label{EqMnAn}
\left[\begin{matrix} M_d & A_d \cr A_d^T & B_d\end{matrix}\right]\left[\begin{matrix} v_d \cr w_d\end{matrix}\right]=\lambda_h(M)\left[\begin{matrix} v_d \cr w_d \end{matrix}\right] \ \Rightarrow \ (M_d-\lambda_h(M)I)v_d=-A_dw_d.
\end{equation}
Let $v_d$ be given in terms of the orthonormal eigenvectors 
$s_1(d),\dots,s_d(d)$ of $M_d$ associated with $\lambda_1(d) \ge \lambda_2(d) \ge \lambda_3(d) \ge \dots$. So
$$
v_d=a_1(d)s_1(d)+\dots+a_d(d)s_d(d)
$$
where $\|a(d)\|_2 \le 1$ since $\|v_d\|_2^2=1-\|w_d\|_2^2 \le 1$. Plugging such expression of $v_d$ 
in (\ref{EqMnAn}) one obtains 
$$
\sum_{i=1}^d \ a_i^2(d)(\lambda_h(M)-\lambda_i(d))^2=\|A_dw_d\|_2^2
$$
with $ \ \|a(d)\|_2 \le 1, \ \|w_d\|_2 \le 1$. 
From (\ref{Eqlemma2eig}), (\ref{Eqlemma3eig})  and the fact that $\|w_d\|_2$ is converging to zero as $d$ goes to $\infty$, one obtains  
that $\|A_dw_d\|_2^2$ is also infinitesimal w.r.t. $d$. So one has 
\begin{subequations} \label{Sumai}
\begin{align}
& \sum_{i=1}^d \ a_i^2(d)(\lambda_h(M)-\lambda_i(d))^2 \ \rightarrow \ 0 \\  
& \sum_{i=1}^d \ a_i^2(d) \ \rightarrow \ 1.
\end{align}
\end{subequations} 
Now, let us assume that the multiplicity of $\lambda_h(M) \ne 0$ is $\nu_h=1$.
The eigenvalues convergence ensures that we can choose $\epsilon >0$ less than an half of the minimum between $\lambda_{h-1}(M)-\lambda_h(M)$ and $\lambda_{h}(M)-\lambda_{h+1}(M)$ such that, for $k$ fixed and $k \ge h$, there exists $N$ such that $d \ge N$ implies $|\lambda_h(M)-\lambda_h(d)| < \epsilon$, while $|\lambda_h(M)-\lambda_j(d)|>\epsilon$ for all $j \ne h$. 
It follows from (\ref{Sumai}) that the $a_i(d)$ with $i \ne h$ decay to zero as $d$ goes to $\infty$. This implies
$$
v_d=a_h(d)s_h(d)+\epsilon_d, \ \text{with} \ \ a_h^2(d) \rightarrow 1, \ \|\epsilon_d\|_2 \rightarrow 0
$$
showing that, as $d$ goes to $\infty$, one has $\| \pm s_h(d)-v_d\|_2 \rightarrow 0$ where $\pm s_h(d)$ 
is the eigenvector (possibly corrected to sign) corresponding to the (only) eigenvalue of $M_d$ which tends to $\lambda_h(M)$.   

\begin{remark}\label{RemarkEigMult}
If the eigenvalue multiplicity is $\nu_h > 1$ for some $h$, the eigenvectors are not well-defined, since there exist infinitely many orthonormal bases for the eigenspace. The $\nu_h-$dimensional eigenspace is approximated (in some sense) by the corresponding space generated by the eigenvectors of $M_d$ corresponding to the eigenvalues which tend to the same $\lambda_h(M)$. However, nothing can be said about the behavior of the single $a_i(d)$, since this is strongly related to the choice of the eigenvectors $s_i(d)$. One has thus to consider $\nu_h$ eigenvectors which tend to lie in a $\nu_h-$dimensional eigenspace.
\end{remark}

\subsection{Proof of Theorem \ref{NecSuffStabKer2}}

Let ${\mathcal M}$ be the operator induced by the kernel $M$, thought of as a map 
from $\ell_{\infty}$ into $\ell_1$. Let $\|{\mathcal M}\|_{\infty,1}$ denote its operator norm.
Then, we know from Theorem \ref{NecSuffStabKer}, and subsequent discussions, that the necessary and sufficient condition
for the stability of the RKHS induced by $M$ is
\begin{equation}\label{OpNormCNS}
\|{\mathcal M}\|_{\infty,1} < +\infty.
\end{equation}
Then, let $(\lambda_i,\rho_i)$ be the eigenvalues and the eigenvectors  orthogonal in $\ell_2$  associated with ${\mathcal M}$. 
The function
$$
f(u) := \|y\|_1=\sum_i \ |y(i) |=\sum_i \ \Big| \sum_h \ M_{ih} u(h) \Big|
$$
is convex being sums of compositions of absolute values and linear functions. 
This permits to state that, for any fixed variable $u(h)$, 
its maximum value is obtained either for $u(h)=+1$ or $u(h)=-1$. By an inductive reasoning we thus obtain
\begin{eqnarray*}
\|{\mathcal M}\|_{\infty,1}&=&\sup_{u \in {\mathcal U}_{\infty}} \ f(u) = \sup_{u \in {\mathcal U}_{\infty}} \ \sum_i \ \Big|\sum_h \ M_{ih}u(h) \Big|.
\end{eqnarray*}
Now, let $M=UDU^T$, where $D$ is diagonal and contains the eigenvalues of $M$
while the columns of $U$ are the corresponding eigenvectors. Then, we have
$$
y=Uw, \ \ w=DU^Tu
$$
and, hence, 
\begin{eqnarray*}
w&=&\left[\begin{matrix} \lambda_1<\rho_1,u>_2 & \ \lambda_2<\rho_2,u>_2 & \dots\end{matrix} \right]^T \\
 y&=&\lambda_1<\rho_1,u>_2\rho_1+\lambda_2<\rho_2,u>_2\rho_2+\dots.
\end{eqnarray*}
To evaluate $\|y\|_1$, we need to consider the scalar product $\langle s(u),y \rangle_2$, where $s(u)=\text{sign}(y)$ (since $y$ depends on $u$, also $s(u)$ does).
In fact, we have
\begin{eqnarray*}
&& h(u):= \|y\|_1 =\sum_h \ \lambda_h<\rho_h,u>_2<\rho_h,s(u)>_2  
\end{eqnarray*}
and this implies
 \begin{eqnarray*}
\|{\mathcal M}\|_{\infty,1}& =&\sup_{u \in {\mathcal U}_{\infty}} \ \sum_h \ \lambda_h<\rho_h,u>_2<\rho_h,s(u)>_2\\
&=& \sup_{u \in {\mathcal U}_{\infty}} \  h(u).
\end{eqnarray*}
Consider also
$$
g(u):=\Sigma_h \ \lambda_h \langle \rho_h,u\rangle_2^2
$$
and define
$$
A:=\sup_{u \in {\mathcal U}_{\infty}} \ \sum_h \ \lambda_h \langle \rho_h,u \rangle_2^2=\sup_{u \in {\mathcal U}_{\infty}} \ g(u).
$$
By definition of $s(u)$, it follows that
$$
h(u) \ge g(u) \ \implies \ \|{\mathcal M}\|_{\infty,1} \ge A.
$$
On the other hand
 \begin{eqnarray*}
h(u)&=&\sum_h \ \lambda_h \langle \rho_h,u\rangle_2 \langle \rho_h,s(u)\rangle_2 \\
&=& \sum_h \ \Big(\sqrt{\lambda_h} \langle \rho_h,u \rangle_2\Big) \ \Big(\sqrt{\lambda_h}\langle_2 \rho_h,s(u)\rangle_2\Big) \\ 
&\le& \sqrt{\sum_h \ \lambda_h \langle \rho_h,u\rangle_2^2}\sqrt{\sum_h \ \lambda_h\langle \rho_h,s(u)\rangle_2^2} \\
&\le& \sqrt{g(u)}\sqrt{g(s(u))}
\end{eqnarray*}
that implies  
$$
\|{\mathcal M}\|_{\infty,1} \le A.
$$
So, one has 
$$
\|{\mathcal M}\|_{\infty,1} = \sup_{u \in {\mathcal U}_{\infty}} \ \sum_h \ \lambda_h \langle \rho_h,u \rangle_2^2.
$$
and this, in view of the necessary and sufficient stability condition (\ref{OpNormCNS}), concludes the proof.

\subsection{Proof of Theorem \ref{NecSuffStabKer3}}

Let again $(\lambda_i,\rho_i)$ be the eigenvalues and the eigenvectors orthogonal in $\ell_2$ associated with 
${\mathcal M}$, the kernel operator induced by $M$. 
One has
 \begin{eqnarray*}
|\langle \rho_h,u \rangle_2| &=& \left|\sum_i \ \rho_h(i)u(i) \right| \le \sum_i \ |\rho_h(i) | |u(i)|  \\
&\le& \sum_i \ |\rho_h(i)| =\| \rho_h \|_1.
\end{eqnarray*}
So, if $\sum_h \ \lambda_h \|\rho_h \|_1^2 < +\infty$,
the above inequality and Theorem \ref{NecSuffStabKer2} ensure stability.\\
In addition, since $M_{ij}=\sum_h \ \lambda_h \rho_h(i) \rho_h(j)$, one has
$$
|M_{ij}| \le \sum_h \ \lambda_h| \rho_h(i)\|\rho_h(j)|=:f_{ij}.
$$
Hence, if $\sum_h \ \lambda_h \|\rho_h \|_1^2 < +\infty$ one obtains
$$
\sum_{ij} \ |M_{ij}| \le \sum_{ij} \ f_{ij}=\sum_h \ \lambda_h\|\rho_h\|_1^2<+\infty
$$
and this proves also the absolute summability of $M$. 

\subsection{Proof of Theorem \ref{NecSuffStabKer4}}

If there exists $A > 0$ such that $\|\rho_h\|_1 \leq A$ if $\lambda_h >0$ and the eigenvalues are summable, one has
$$
\sum_h \ \lambda_h \|\rho_h\|_1^2 \leq A^2 \sum_h  \lambda_h < +\infty 
$$
and Theorem \ref{NecSuffStabKer3} then ensures stability.\\
If the kernel $M$ is stable, its trace is finite and from Lemma \ref{ThStableCompact} we know that the kernel operator 
$\mathcal{M}$ is compact. Then, as discussed during the proof of Theorem \ref{ThRKHSinclusions}, it holds that
$$
\text{tr}(M)= \sum_h  \lambda_h <+\infty
$$
and this concludes the proof.


\begin{thebibliography}{10}

\bibitem{ArgyriouD2014}
A.~Argyriou and F.~Dinuzzo.
\newblock A unifying view of representer theorems.
\newblock In {\em Proceedings of the 31th International Conference on Machine
  Learning}, volume~32, pages 748--756, 2014.

\bibitem{Aronszajn50}
N.~Aronszajn.
\newblock Theory of reproducing kernels.
\newblock {\em Transactions of the American Mathematical Society}, 68:337--404,
  1950.

\bibitem{Atkinson1975}
K.~Atkinson.
\newblock Convergence rates for approximate eigenvalues of compact integral
  operators.
\newblock {\em SIAM Journal on Numerical Analysis}, 12(2):213--222, 1975.

\bibitem{Baker1977}
C.~Baker.
\newblock {\em The numerical treatment of integral equations}.
\newblock Clarendon press, 1977.

\bibitem{Mik2018}
M.~Belkin, S.~Ma, and S.~Mandal.
\newblock {To understand deep learning we need to understand kernel learning}.
\newblock {\em arXiv e-prints}, Feb 2018.

\bibitem{Bergman50}
S.~Bergman.
\newblock {\em The Kernel Function and Conformal Mapping}.
\newblock Mathematical Surveys and Monographs, AMS, 1950.

\bibitem{Bertero:1988}
M.~Bertero, T.~Poggio, and V.~Torre.
\newblock Ill-posed problems in early vision.
\newblock In {\em Proceedings of the IEEE}, pages 869--889, 1988.

\bibitem{SiamArxivAKS2019}
M.~Bisiacco and G.~Pillonetto.
\newblock Kernel absolute summability is only sufficient for {RKHS} stability.
\newblock {\em {A}r{X}iv e-prints 1909.02341 2019
  \url{https://arxiv.org/abs/1909.02341}}.

\bibitem{BAHP16}
G.~Bottegal, A.Y. Aravkin, H.~Hjalmarsson, and G.~Pillonetto.
\newblock Robust {EM} kernel-based methods for linear system identification.
\newblock {\em Automatica}, 67:114 -- 126, 2016.

\bibitem{CarliIFAC12}
F.P. Carli, A.~Chiuso, and G.~Pillonetto.
\newblock Efficient algorithms for large scale linear system identification
  using stable spline estimators.
\newblock In {\em Proceedings of the 16th IFAC Symposium on System
  Identification (SysId 2012)}, 2012.

\bibitem{Carmeli}
C.~Carmeli, E.~De Vito, and A.~Toigo.
\newblock {V}ector valued reproducing kernel {H}ilbert spaces of integrable
  functions and {M}ercer theorem.
\newblock {\em Analysis and Applications}, 4:377--408, 2006.

\bibitem{ChenKS2018}
T.~Chen.
\newblock On kernel design for regularized lti system identification.
\newblock {\em Automatica}, 90:109 -- 122, 2018.

\bibitem{ChenetalTAC:13}
T.~Chen, M.~S. Andersen, L.~Ljung, A.~Chiuso, and G.~Pillonetto.
\newblock System identification via sparse multiple kernel-based regularization
  using sequential convex optimization techniques.
\newblock {\em IEEE Transactons on Automatic Control}, provisionally accepted,
  2013.

\bibitem{ChenKS2015}
T.~Chen and L.~Ljung.
\newblock On kernel structures for regularized system identification (ii): a
  system theory perspective.
\newblock {\em IFAC-PapersOnLine}, 48(28):1041 -- 1046, 2015.
\newblock 17th IFAC Symposium on System Identification SYSID 2015.

\bibitem{ChenOrth2015}
T.~{Chen} and L.~{Ljung}.
\newblock Regularized system identification using orthonormal basis functions.
\newblock In {\em 2015 European Control Conference (ECC)}, pages 1291--1296,
  2015.

\bibitem{mkcdc12}
T.~Chen, L.~Ljung, M.~Andersen, A.~Chiuso, F.P. Carli, and G.~Pillonetto.
\newblock Sparse multiple kernels for impulse response estimation with
  majorization minimization algorithms.
\newblock In {\em IEEE Conference on Decision and Control}, pages 1500--1505,
  Hawaii, Dec 2012.

\bibitem{ChenOL12}
T.~Chen, H.~Ohlsson, and L.~Ljung.
\newblock On the estimation of transfer functions, regularizations and
  {G}aussian processes - revisited.
\newblock {\em Automatica}, 48(8):1525--1535, 2012.

\bibitem{ChenStableRKHS}
T.~Chen and G.~Pillonetto.
\newblock On the stability of reproducing kernel {H}ilbert spaces of
  discrete-time impulse responses.
\newblock {\em Automatica}, 95:529 -- 533, 2018.

\bibitem{SSvsNN2013}
A.~Chiuso, T.~Chen, L.~Ljung, and G.~Pillonetto.
\newblock Regularization strategies for nonparametric system identification.
\newblock In {\em Proceedings of the 52nd Annual Conference on Decision and
  Control (CDC)}, 2013.

\bibitem{KBdeep2009}
Y.~Cho and L.K. Saul.
\newblock Kernel methods for deep learning.
\newblock In Y.~Bengio, D.~Schuurmans, J.~D. Lafferty, C.~K.~I. Williams, and
  A.~Culotta, editors, {\em Advances in Neural Information Processing Systems
  22}, pages 342--350. Curran Associates, Inc., 2009.

\bibitem{Cucker01}
F.~Cucker and S.~Smale.
\newblock On the mathematical foundations of learning.
\newblock {\em Bulletin of the American mathematical society}, 39:1--49, 2001.

\bibitem{Darwish2015}
M.~{Darwish}, G.~{Pillonetto}, and R.~{Tóth}.
\newblock Perspectives of orthonormal basis functions based kernels in bayesian
  system identification.
\newblock In {\em 2015 54th IEEE Conference on Decision and Control (CDC)},
  pages 2713--2718, 2015.

\bibitem{DARWISH2018318}
M.A.H. Darwish, G.~Pillonetto, and R.~Toth.
\newblock The quest for the right kernel in bayesian impulse response
  identification: The use of obfs.
\newblock {\em Automatica}, 87:318 -- 329, 2018.

\bibitem{DinuzzoSIAM15}
F.~Dinuzzo.
\newblock Kernels for linear time invariant system identification.
\newblock {\em SIAM Journal on Control and Optimization}, 53(5):3299--3317,
  2015.

\bibitem{Drucker97}
H.~Drucker, C.J.C. Burges, L.~Kaufman, A.~Smola, and V.~Vapnik.
\newblock Support vector regression machines.
\newblock In {\em Advances in Neural Information Processing Systems}, 1997.

\bibitem{Dunford1963}
N.~Dunford and J.T. Schwartz.
\newblock {\em Linear operators}.
\newblock InterScience Publishers, 1963.

\bibitem{Fujimoto2017}
Y.~Fujimoto, I.~Maruta, and T.~Sugie.
\newblock Extension of first-order stable spline kernel to encode relative
  degree.
\newblock {\em IFAC-PapersOnLine}, 50(1):14016 -- 14021, 2017.
\newblock 20th IFAC World Congress.

\bibitem{Girosi:1997}
F.~Girosi.
\newblock An equivalence between sparse approximation and support vector
  machines.
\newblock Technical report, Cambridge, MA, USA, 1997.

\bibitem{Gittens:2016}
A.~Gittens and M.~Mahoney.
\newblock Revisiting the nystr\"{o}m method for improved large-scale machine
  learning.
\newblock {\em J. Mach. Learn. Res.}, 17(1):3977--4041, 2016.

\bibitem{OrthBook2005}
P.~Heuberger, P.~{van den} Hof, and B.~Wahlberg.
\newblock {\em Modelling and Identification with Rational Orthogonal Basis
  Functions}.
\newblock Springer, 2005.

\bibitem{Kimeldorf70}
G.~Kimeldorf and G.~Wahba.
\newblock A correspondence between bayesian estimation on stochastic processes
  and smoothing by splines.
\newblock {\em The Annals of Mathematical Statistics}, 41(2):495--502, 1970.

\bibitem{Kumar:2012}
S.~Kumar, M.~Mohri, and A.~Talwalkar.
\newblock Sampling methods for the {N}ystr\"{o}m method.
\newblock {\em J. Mach. Learn. Res.}, 13(1):981--1006, 2012.

\bibitem{Lidskii1959}
V.B. Lidskii.
\newblock Non-self-adjoint operators with a trace.
\newblock {\em Dokl. Akad. Nauk.}, 1959.

\bibitem{Ljung:99}
L.~Ljung.
\newblock {\em System Identification - Theory for the User}.
\newblock Prentice-Hall, Upper Saddle River, N.J., 2nd edition, 1999.

\bibitem{LCB2019}
L.~Ljung, T.~Chen, and B.~Mu.
\newblock A shift in paradigm for system identification.
\newblock {\em International Journal of Control}, pages 1--8, 2019.

\bibitem{Minh2010}
H.Q. Minh.
\newblock Some properties of {G}aussian reproducing kernel {H}ilbert spaces and
  their implications for function approximation and learning theory.
\newblock {\em Constr. Approx.}, 32(2):307--338, 2010.

\bibitem{NinnessOrth1999}
B.~{Ninness}, H.~{Hjalmarsson}, and F.~{Gustafsson}.
\newblock The fundamental role of general orthonormal bases in system
  identification.
\newblock {\em IEEE Transactions on Automatic Control}, 44(7):1384--1406, 1999.

\bibitem{PillonettoWien13}
G.~Pillonetto.
\newblock Consistent identification of {W}iener systems: A machine learning
  viewpoint.
\newblock {\em Automatica}, 49(9):2704--2712, September 2013.

\bibitem{PillonettoHybrid}
G.~Pillonetto.
\newblock A new kernel-based approach to hybrid system identification.
\newblock {\em Automatica}, 70:21 -- 31, 2016.

\bibitem{PillInsights2018}
G.~Pillonetto.
\newblock System identification using kernel-based regularization: New insights
  on stability and consistency issues.
\newblock {\em Automatica}, 93:321--332, 2018.

\bibitem{PilAuto2007}
G.~Pillonetto and B.M. Bell.
\newblock Bayes and empirical {B}ayes semi-blind deconvolution using
  eigenfunctions of a prior covariance.
\newblock {\em Automatica}, 43(10):1698--1712, 2007.

\bibitem{Pillonetto2016}
G.~Pillonetto, T.~Chen, A.~Chiuso, G.~{De Nicolao}, and L.~Ljung.
\newblock Regularized linear system identification using atomic, nuclear and
  kernel-based norms: The role of the stability constraint.
\newblock {\em Automatica}, 69:137 -- 149, 2016.

\bibitem{PillonettoMLrob2015}
G.~Pillonetto and A.~Chiuso.
\newblock Tuning complexity in regularized kernel-based regression and linear
  system identification: The robustness of the marginal likelihood estimator.
\newblock {\em Automatica}, 58:106 -- 117, 2015.

\bibitem{PillACC2010}
G.~Pillonetto, A.~Chiuso, and G.~{De Nicolao}.
\newblock Regularized estimation of sums of exponentials in spaces generated by
  stable spline kernels.
\newblock In {\em Proceedings of the IEEE American Cont. Conf., Baltimora,
  USA}, 2010.

\bibitem{SS2010}
G.~Pillonetto and G.~{De Nicolao}.
\newblock A new kernel-based approach for linear system identification.
\newblock {\em Automatica}, 46(1):81--93, 2010.

\bibitem{SurveyKBsysid}
G.~Pillonetto, F.~Dinuzzo, T.~Chen, G.~De Nicolao, and L.~Ljung.
\newblock Kernel methods in system identification, machine learning and
  function estimation: a survey.
\newblock {\em Automatica}, 50(3):657--682, 2014.

\bibitem{PillPAMI2}
G.~Pillonetto, L.~Schenato, and D.~Varagnolo.
\newblock Distributed multi-agent {G}aussian regression via finite-dimensional
  approximations.
\newblock {\em IEEE Trans. on Pattern Analysis and Machine Intelligence},
  41(9):2098--2111, 2019.

\bibitem{Poggio90}
T.~Poggio and F.~Girosi.
\newblock {N}etworks for approximation and learning.
\newblock In {\em Proceedings of the {IEEE}}, volume~78, pages 1481--1497,
  1990.

\bibitem{Robert2017}
D.~Robert.
\newblock On the traces of operators (from {G}rothendieck to {L}idskii).
\newblock {\em EMS newsletter}, 2017.

\bibitem{Rudin}
W.~Rudin.
\newblock {\em Real and Complex Analysis}.
\newblock McGraw-Hill, Singapore, 1987.

\bibitem{Scholkopf01}
B.~Sch\"{o}lkopf, R.~Herbrich, and A.~J. Smola.
\newblock A generalized representer theorem.
\newblock {\em Neural Networks and Computational Learning Theory}, 81:416--426,
  2001.

\bibitem{Sun05}
Hongwei Sun.
\newblock Mercer theorem for {RKHS} on noncompact sets.
\newblock {\em J. Complexity}, 21(3):337--349, 2005.

\bibitem{Vapnik98}
V.~Vapnik.
\newblock {\em Statistical Learning Theory}.
\newblock Wiley, New York, NY, USA, 1998.

\bibitem{Wahba1990}
G~Wahba.
\newblock {\em {Spline Models For Observational Data}}.
\newblock SIAM, Philadelphia, 1990.

\bibitem{WalLag1991}
B.~{Wahlberg}.
\newblock System identification using {L}aguerre models.
\newblock {\em IEEE Transactions on Automatic Control}, 36(5):551--562, 1991.

\bibitem{WalIFAC1994}
B.~Wahlberg.
\newblock Laguerre and {K}autz models.
\newblock {\em IFAC Proceedings Volumes}, 27(8):965 -- 976, 1994.
\newblock IFAC Symposium on System Identification (SYSID'94), Copenhagen,
  Denmark, 4-6 July.

\bibitem{Williams:2000}
C.K.~I. Williams and M.~Seeger.
\newblock Using the {N}ystr\"{o}m method to speed up kernel machines.
\newblock In {\em Proceedings of the 2000 conference on Advances in neural
  information processing systems}, page 682Ð688, Cambridge, MA, USA, 2000. MIT
  Press.

\bibitem{Zeidler}
E.~Zeidler.
\newblock {\em Applied Functional Analysis}.
\newblock Springer, 1995.

\bibitem{Zhu98}
H.~Zhu, C.K.I. Williams, R.J. Rohwer, and M.~Morciniec.
\newblock Gaussian regression and optimal finite dimensional linear models.
\newblock In C.~M. Bishop, editor, {\em Neural Networks and Machine Learning}.
  Springer-Verlag, Berlin, 1998.

\end{thebibliography}

\end{document}